\RequirePackage{fix-cm}
\documentclass[twocolumn]{svjour3}          

\smartqed  

\usepackage{graphicx}
\usepackage{amsmath}
\usepackage{amssymb}
\usepackage{times}
\usepackage{amsmath}
\usepackage{amssymb}
\usepackage{array}
\usepackage{url}
\usepackage{comment}
\usepackage{rotating}
\usepackage{multirow}
\usepackage{subfigure}
\usepackage{color}
\usepackage{acronym}
\usepackage{algorithmic}
\usepackage[normalem]{ulem}
\usepackage{refcount}
\usepackage[pagebackref=true,breaklinks=true,letterpaper=true,colorlinks,bookmarks=false]{hyperref}
\usepackage{enumitem}
\usepackage[toc,page]{appendix}
\usepackage{xspace}
\usepackage{pbox}

\makeatletter
\DeclareRobustCommand\onedot{\futurelet\@let@token\bmv@onedotaux}
\def\bmv@onedotaux{\ifx\@let@token.\else.\null\fi\xspace}
\def\eg{\emph{e.g}\onedot}

\def\etc{\emph{etc}\onedot} 
 
\def\etal{\emph{et al}\onedot}

\makeatother

\newcommand{\mI}{{\boldsymbol{I}}}
\newcommand{\mIh}{{\widehat{\boldsymbol{I}}}}
\newcommand{\mTheta}{{\boldsymbol{\Theta}}}
\newcommand{\mPhi}{{\boldsymbol{\Phi}}}
\newcommand{\vpsi}{{\boldsymbol{\psi}}}
\newcommand{\mG}{{\boldsymbol{G}}}
\newcommand{\mD}{{\boldsymbol{D}}}

\begin{document}

\title{Identity-preserving Face Recovery from Stylized Portraits}

\author{Fatemeh~Shiri\textsuperscript{1} \and Xin~Yu\textsuperscript{1} \and Fatih~Porikli\textsuperscript{1} \and Richard~Hartley\textsuperscript{1,2} \and Piotr~Koniusz\textsuperscript{2,1} 
}


\institute{F. Shiri\textsuperscript{1}, X.Yu\textsuperscript{1}, F. Porikli\textsuperscript{1}, R.Hartley\textsuperscript{1,2}, P. Koniusz\textsuperscript{2,1}\\\vspace{-0.2cm}\at
1-Australian National University,
\email{name.surname@anu.edu.au}           
\and
2-Data61/CSIRO, \email{name.surname@data61.csiro.au}
}

\date{Received: 23.02.2018 / Accepted: 29.01.20191}

\maketitle

\begin{abstract}

Given an artistic portrait, recovering the latent photorealistic face that preserves the subject's identity is challenging because the facial details are often distorted or fully lost in artistic portraits. We develop an Identity-preserving Face Recovery from Portraits (IFRP) method that utilizes a Style Removal network (SRN) and a Discriminative Network (DN). Our SRN, composed of an autoencoder with residual block-embedded skip connections, is designed to transfer feature maps of stylized images to the feature maps of the corresponding photorealistic faces. Owing to the Spatial Transformer Network (STN), SRN automatically compensates for misalignments of stylized portraits to output aligned realistic face images. To ensure the identity preservation, we promote the recovered and ground truth faces to share similar visual features via a distance measure which compares features of recovered and ground truth faces extracted from a pre-trained FaceNet network. DN has multiple convolutional and fully-connected layers, and its role is to enforce recovered faces to be similar to authentic faces. 
Thus, we can recover high-quality photorealistic faces from unaligned portraits while preserving the identity of the face in an image. By conducting extensive evaluations on a large-scale synthesized dataset and a hand-drawn sketch dataset, we demonstrate that our method achieves superior face recovery and attains state-of-the-art results. In addition, our method can recover photorealistic faces from unseen stylized portraits, artistic paintings, and hand-drawn sketches. 
\keywords{ Face Synthesis \and Image Stylization \and Face Recovery \and Destylization \and Generative Models}

\end{abstract}

\begin{figure}[t]
\centering
\subfigure[Original]{\label{fig:opena}\scalebox{1}[1]{\includegraphics[width=0.24\linewidth]{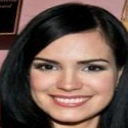}}}
\centering
\subfigure[Portrait]{\label{fig:openb}\scalebox{1}[1]{\includegraphics[width=0.24\linewidth]{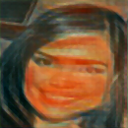}}}
\subfigure[{Landmarks}]{\label{fig:openc}\scalebox{1}[1]{\includegraphics[width=0.24\linewidth]{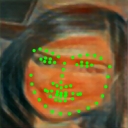}}}
\subfigure[\scriptsize{\cite{johnson2016perceptual}}]{\label{fig:opencc}\scalebox{1}[1]{\includegraphics[width=0.24\linewidth]{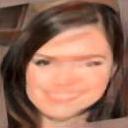}}}\\
\subfigure[\scriptsize{\cite{zhu2017unpaired}}]{\label{fig:openccc}\scalebox{1}[1]{\includegraphics[width=0.24\linewidth]{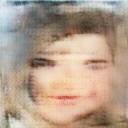}}}
\subfigure[\cite{isola2016image}$\!\!$]{\label{fig:openf}\scalebox{1}[1]{\includegraphics[width=0.24\linewidth]{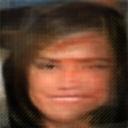}}}
\subfigure[~\cite{Shiri2017FaceD}]{\label{fig:opene}\scalebox{1}[1]{\includegraphics[width=0.24\linewidth]{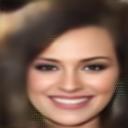}}}
\subfigure[Ours]{\label{fig:openg}\scalebox{1}[1]{\includegraphics[width=0.24\linewidth]{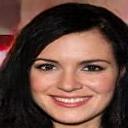}}}
\caption{Comparisons to the state-of-art methods. (a) The ground truth face image (from test dataset, not available in the training dataset). (b) Unaligned stylized portraits of (a) from \emph{Scream} style. (c) Landmarks detected  by~\cite{zhang2014facial}. (d) Results obtained by ~\cite{johnson2016perceptual}. (e) Results obtained by~\cite{zhu2017unpaired} (CycleGAN). (f) Results obtained by ~\cite{isola2016image} (pix2pix). (g) Results obtained by~\cite{Shiri2017FaceD}. (h) Our results.} 
\label{fig:open}
\vspace{-0.4cm}
\end{figure}

\section{Introduction}
\label{intro}
Style transferring methods are powerful tools that can generate portraits in various artistic styles from photorealistic images.
Unlike prior research on the image stylization, we address a challenging inverse problem of {\em photorealistic face recovery from stylized portraits} which aims at recovering a photorealistic image of face from a given stylized portrait.
Latent photorealistic face images recovered from their artistic portraits are interpretable for humans and they may be useful in facial analysis.
Since facial details and expressions in stylized portraits often undergo severe distortions and become corrupted by artifacts such as profile edges and color changes \eg, as in Figure \ref{fig:openb}, recovering a photorealistic face image from its stylized counterpart is very challenging. In general, stylized face images contain a variety of facial expressions, facial feature distortions and misalignments. Therefore, landmark detectors often fail to localize facial landmarks accurately as shown in Figure \ref{fig:openc}.   

While recovering photorealistic images from portraits is still uncommon in the literature, image stylization methods have been widely studied. With the use of Convolutional Neural Networks (CNN), Gatys~\emph{et al.}~\cite{gatys2016controlling} achieve promising results by transferring different styles of artworks to images via the semantic contents space. Since their method generates the stylized images by iteratively updating the feature maps of CNNs, it is computationally costly. In order to reduce the computational complexity, several feed-forward CNN-based methods have been proposed \cite{ulyanov2016texture,ulyanov2016instance,johnson2016perceptual,dumoulin2016,li2017diversified,chen2016fast,zhang2017multi,huang2017arbitrary}. 
However, these methods work only with a single style applied during training. Moreover, such methods are insufficient for generating photorealistic face images as they only capture the correlations of feature maps via Gram matrices thus  discarding spatial relations \cite{pk_tensor,me_museum,power_look_cvpr}.

In order to capture spatial/localized statistics of a style image, several patch-based methods~\cite{li2016precomputed,isola2016image} have been developed. However, such methods cannot capture the global appearance of faces, thus failing to generate authentic face images. For instance, patch-based methods \cite{li2016precomputed,isola2016image} fail to attain consistency of face colors, as shown in Figure~\ref{fig:cmp2e}. Moreover, the state-of-the-art style transfer methods~\cite{gatys2016controlling,li2016precomputed,ulyanov2016texture,johnson2016perceptual} transfer  desired styles to images without considering the task of identity preservation. Thus, these methods cannot generate realistically looking faces with preserved identities.

Our first face destylization architecture \cite{Shiri2017FaceD} uses only a pixel-wise loss in the generative part of the network. Despite being trained on a large-scale dataset, this method fails to recover faces from unaligned portraits under a variety of scales, rotations and viewpoint variations. This journal manuscript is an extension of our second model \cite{Shiri2018wacv} which introduces the  identity-preserving loss into destylization. Our latest model \cite{Shiri2019wacv} performs an identity-preserving face destylization with the use of attributes which allow to manipulate appearance details such as hair color, facial expressions, \etc.

In this paper, we develop a novel end-to-end trainable identity-preserving approach to face recovery that automatically maps the unaligned stylized portraits to aligned photorealistic face images. 
Our network employs two subnetworks: a generative subnetwork, dubbed Style Removal Network (SRN), and a Discriminative Network (DN). 
SRN consists of an autoencoder (a downsampling encoder and an upsampling decoder) and Spatial Transfer Networks (STN)~\cite{jaderberg2015spatial}.
The encoder extracts facial components from unaligned stylized face images to transfer the extracted feature maps to the domain of photorealistic images. Subsequently, our decoder forms face images. STN layers are used by the encoder and decoder to align stylized faces. 
Since faces may appear at different orientations, scales and in various poses, the network may not fully capture all this variability if the  training data does not account for it. As a result, we would need heavy data augmentation and more training instances with variety of poses in the training dataset to cope with recovery of faces from authentic portraits that may be presented under angle or viewpoint, \etc. In contrast to such a costly training, by exploiting STN layers, we require less data to train our network to cope well with images containing face rotations, translations and scale changes. Nonetheless, with or without STN layers, we expose our network during training to images of faces at different scales and rotations to train it how to recover the frontal view. 
We aim to recover faces in frontal view for visualization purposes (easy to interpret for humans, a face retrieval software works better with frontal views, \etc).
The discriminative network, inspired by approaches~\cite{Goodfellow2014,denton2015deep,yu2016ultra,yu2017face}, forces SRN to generate destylized faces to be similar to authentic ground truth faces.

As we aim to preserve the information about facial identities, we force the CNN feature representations of recovered faces to be as close to the features of ground truth real faces as possible. For this purpose, we employ pixel-level Euclidean and identity-preserving losses. 
We also use an adversarial loss to achieve  high-quality visual results.   

To train our network, pairs of Stylized Face (SF) and ground truth Real Face (RF) images are required. Thus, we synthesize a large-scale dataset of SF/RF pairs. 
%
As there exist numerous styles to choose from, we cannot generate faces in all possible styles for training. We note that a Gram matrix formed from features of pre-trained VGG network can capture style details of input images~\cite{gatys2016image}. Thus, we measure the similarity of various styles via the Log-Euclidean distance~\cite{jayasumana2013kernel} between Gram matrices of style images and the average Gram matrix of real faces. 
Based on such a style-distance metric, we select three distinct styles for training.

Moreover, we have observed that CNN filters learned on images of seen styles (used for training) tend to extract meaningful features from images in both seen and unseen styles. Thus, our method can also extract facial information from unseen stylized portraits and generate photorealistic faces, as demonstrated in the experimental section.

Below we list our contributions:
\renewcommand{\labelenumi}{\Roman{enumi}.}
\begin{enumerate}
\vspace{-2mm}
\item We design a new framework to automatically remove styles from unaligned stylized portraits. Our approach generates facial identities and expressions that match the ground truth face images well (identity preservation). 
\item We propose an autoencoder with skip connections between top convolutional and deconvolutional layers; each skip connection being composed of three residual blocks. These skip connections pass high-level visual features of portraits from convolutional to deconvolutional layers, which leads to an improved restoration performance. 

\item We add an identity-preserving loss to remove seen/unseen styles from portraits preserve underlying identities.

\item We use STNs as intermediate layers to learn to align non-aligned input portraits. Thus, our method does not use any facial landmarks or 3D models of faces (typically used for face alignment) and requires somewhat fewer augmentations than a network without STNs.

\item We propose a style-distance metric to capture the most distinct styles for training. Thus, our network achieves a good generalization when tested on unseen styles. 
\end{enumerate}

Our large dataset of pairs of stylized and photorealistic faces, and the code will be available on {\fontsize{9}{9}\selectfont \url{https://github.com/fatimashiri}} and/or {\fontsize{9}{9}\selectfont \url{http://claret.wikidot.com}}.

\begin{figure*}[!t]
\centering
\scalebox{1}[1]{\includegraphics[width=1\linewidth]{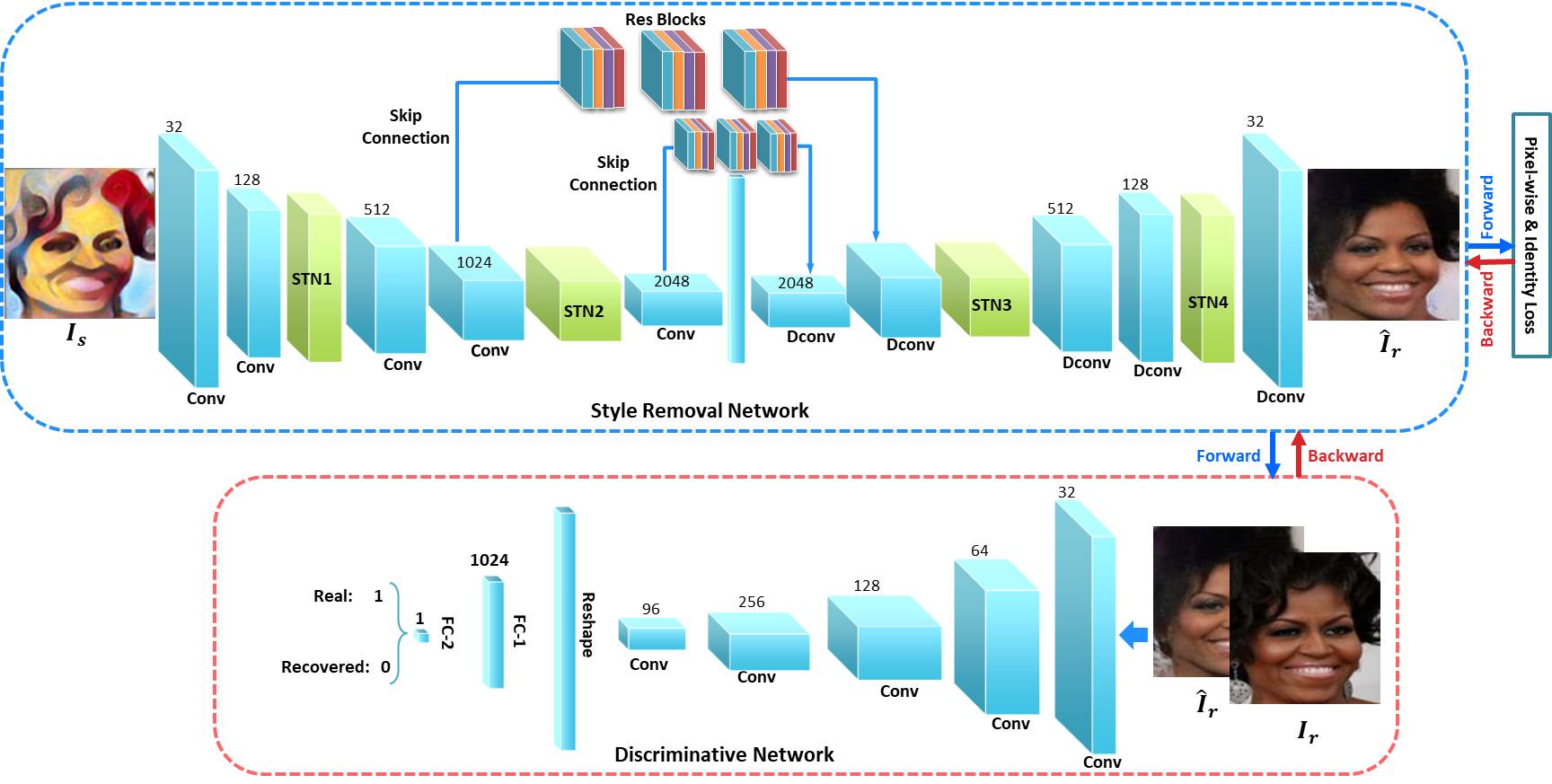}}
\caption{Our identity-preserving face destylization framework consists of two parts: a style removal network (blue frame) and a discriminative network (red frame). The face recovery network takes portraits as inputs. The discriminative network takes real or recovered face images as inputs.}
\label{fig:pipeline}
\end{figure*}

\section{Related Work}
\label{sec:Related Work}
In this section, we briefly review neural generative models and deep style transfer methods for image generation.

\subsection{Neural Generative Models}
There exist many generative models for the problem of image generation \cite{oord2016pixel,kingma2013auto,oord2016pixel,Goodfellow2014,denton2015deep,zhang2017image,Shiri2017FaceD}. Among them, GANs are conceptually closely related to our problem as they employ an adversarial loss that forces the generated images to be as photorealistic as the ground truth images. 

Several methods for super-resolution \cite{ledig2016photo,yu2017face,huang2017beyond,yu2017hallucinating,yu2016ultra} and inpainting~\cite{pathak2016context} adopt an adversarial training to learn a parametric translating function from a large-scale dataset of input-output pairs. These approaches often use the $\ell_1$ or $\ell_2$ norm and adversarial losses to compare the generated image to the corresponding ground truth image. Although these methods produce impressive photorealistic images, they fail to preserve identities of subjects.

Conditional GANs have been used for the task of generating photographs from semantic layout/scene attributes \cite{karacan2016learning} and sketches \cite{sangkloy2016scribbler}. Li and Wand \cite{li2016precomputed} train a Markovian GAN for the style transfer -- a discriminative training is applied on Markovian neural patches to capture local style statistics.  
Isola \emph{et al.} \cite{isola2016image} develop ``pix2pix'' framework which uses so-called ``Unet'' architecture and the patch-GAN to transfer low-level features from the input to the output domain. For faces, this approach produces visual artifacts and fails to capture the global appearance of faces.

Patch-based methods fail to capture the global appearance of faces and, as a result, they generate poorly destylized images. 
In contrast, we propose an identity-preserving loss to faithfully recover the most prominent details of faces.

Moreover, there exist several deep learning methods that synthesize sketches from photographs (and vice versa) \cite{nejati2011study,wang2018back,wang2018high,sharma2011bypassing}. 
Wang \emph{et al.}~\cite{wang2018back} use the vanilla conditional GAN (cGAN) to generate sketches. However, the cGAN produces sketch-like artifacts in the synthesized faces as well as facial deformations. Wang \emph{et al.}~\cite{wang2018high} use the CycleGAN~\cite{CycleGAN2017}, and employ multi-scale discriminators to generate high resolution sketches/photos. Their method demonstrates a greatly improved performance. However, it still produces slight blur and/or color degraded artifacts. Kazemi \emph{et al.}~\cite{kazemi2018facial} employ Cycle-GAN conditioned on facial attributes in order to enforce desired facial attributes over the images synthesized from sketches.
While sketch-to-face synthesis is a related problem, our unified framework works well with a variety of styles more complex than sketches.

\subsection{Deep Style Transfer}
Style transfer is a technique which can render a given content image (input) according to a specific painting style while preserving the visual contents of the input. 
We distinguish \emph{image optimization} and \emph{feed-forward} style transfer methods. The seminal optimization-based work~\cite{gatys2016image} transfers the style of an artistic image to a given photograph. It uses iterative optimization to generate a target image from a random initialization (following the Normal distribution). During the optimization step, the statistics of the feature maps of the target, the content and style images are matched.

Gatys~\emph{et al.}~\cite{gatys2016image} inspired many follow-up studies. 
Yin \cite{yin2016content} presents a content-aware style transfer method which initializes the optimization step with a content image instead of a random noise. Li and Wand \cite{li2016combining} propose a patch-based style transfer method which combines Markov Random Field (MRF) and CNN techniques. Gatys~\emph{et al.} \cite{gatys2016preserving} transfer the style via linear models and preserve colors of content images by matching color histograms. 

Gatys~\emph{et al.}~\cite{gatys2016controlling} decompose styles into perceptual factors and then manipulate them for the style transfer. Selim~\emph{et al.} \cite{selim2016painting} modify the content loss through a gain map for the transfer of paintings of head. 
Wilmot ~\emph{et al.}~\cite{wilmot2017stable} use histogram-based losses in their objective and build on the Gatys~\emph{et al.}'s algorithm~\cite{gatys2016image}. Although the above optimization-based methods further improve the quality of style transfer, they are computationally expensive due to the iterative optimization procedure, thus limiting their practical use.

To address the poor computational speed, feed-forward methods replace the original on-line iterative optimization step with training a feed-forward neural network off-line and generating stylized images on-line~\cite{ulyanov2016texture,johnson2016perceptual,li2016precomputed}.

Johnson \emph{et al.} \cite{johnson2016perceptual} train a generative network for a fast style transfer using perceptual loss functions. 
The architecture of their generator network follows the work of \cite{radford2015unsupervised} and also uses residual blocks. Texture Network ~\cite{ulyanov2016texture} employs a multi-resolution architecture in the generator network. 
Ulyanov \emph{et al.} \cite{ulyanov2016instance,ulyanov2017improved} replace the spatial batch normalization with the instance normalization to achieve a faster convergence. Wang \emph{et al.}~\cite{wang2016multimodal} enhance the granularity of the feed-forward style transfer with a multimodal CNN, which performs stylization hierarchically using multiple losses deployed across multiple scales. 

These feed-forward methods perform stylization  around 1000$\times$ faster than the optimization-based methods. However, they cannot adapt to arbitrary styles not used during training. In order to synthesize an image according to a new style, the entire network needs retraining. To deal with such a restriction, a number of recent approaches encode multiple styles within a single feed-forward network \cite{dumoulin2016,chen2016fast,chen2017stylebank,li2017diversified}.

Dumoulin \emph{et al.} \cite{dumoulin2016} use a so-called conditional instance normalization that learns normalization parameters for each style. Given feature maps of the content and style images, method \cite{chen2016fast} replaces content features with the closest matching style features patch-by-patch. Chen \emph{et al.}~\cite{chen2017stylebank} present a network that learns a set of new filters for every new style.   
Li~\emph{et al.}~\cite{li2017diversified} 
propose a texture controller which forces the network to synthesize the desired style. We note that the existing feed-forward approaches have to compromise between the generalization \cite{li2017diversified,huang2017arbitrary,zhang2017multi} and quality \cite{ulyanov2017improved,ulyanov2016instance,gupta2017characterizing}.


\begin{figure}[t]
\centering
\includegraphics[width=0.18\linewidth]{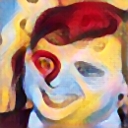}
\includegraphics[width=0.18\linewidth]{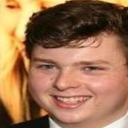}
\includegraphics[width=0.18\linewidth]{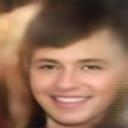}
\includegraphics[width=0.18\linewidth]{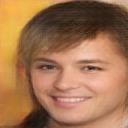}
\includegraphics[width=0.18\linewidth]{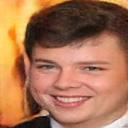}\\
\includegraphics[width=0.18\linewidth]{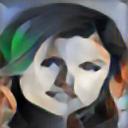}
\includegraphics[width=0.18\linewidth]{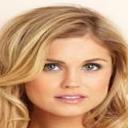}
\includegraphics[width=0.18\linewidth]{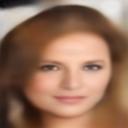}
\includegraphics[width=0.18\linewidth]{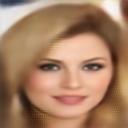}
\includegraphics[width=0.18\linewidth]{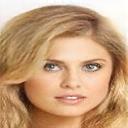}\\
\includegraphics[width=0.18\linewidth]{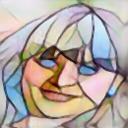}
\includegraphics[width=0.18\linewidth]{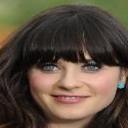}
\includegraphics[width=0.18\linewidth]{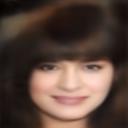}
\includegraphics[width=0.18\linewidth]{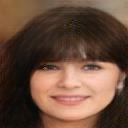}
\includegraphics[width=0.18\linewidth]{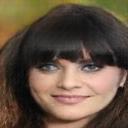}\\
\subfigure[]{\label{fig:DisA}\scalebox{1}[1]{\includegraphics[width=0.18\linewidth]{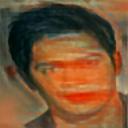}}}
\subfigure[]{\label{fig:DisB}\scalebox{1}[1]{\includegraphics[width=0.18\linewidth]{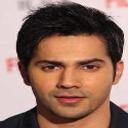}}}
\subfigure[]{\label{fig:DisC}\scalebox{1}[1]{\includegraphics[width=0.18\linewidth]{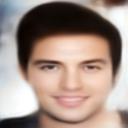}}}
\subfigure[]{\label{fig:DisD}\scalebox{1}[1]{\includegraphics[width=0.18\linewidth]{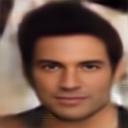}}}
\subfigure[]{\label{fig:DisE}\scalebox{1}[1]{\includegraphics[width=0.18\linewidth]{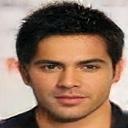}}}
\caption{The contribution of each loss function to IFRP network. (a) Unaligned input portraits from the test dataset. (b) Ground truth face images. (c) Recovered faces; only the pixel-wise loss is used (no DN or identity-preserving losses). (d) Recovered faces; the pixel-wise loss and discriminative loss are used (no identity-preserving loss). (e) Our final results with the pixel-wise, discriminative and identity-preserving losses. The use of all three losses produces visually the best results.}
\label{fig:DiscEffect}  
\end{figure}

\section{Proposed Method} 
Below we present an identity-preserving framework that infers a photorealistic face image $\mIh_r$ from an unaligned stylized face image $\mI_s$.
 
\subsection{Network Architecture}
Our network consists of two parts: a Style Removal Network (SRN) and a Discriminative Network (DN). SRN is composed of an autoencoder as well as skip connections with residual blocks. 
The SRN module extracts residual feature maps from input portraits and then upsamples them. To attain high-quality visual performance, we pass visual information from last few layers of encoder to the corresponding layers of decoder.
The role of DN is to promote the recovered face images to be similar to their real counterparts. 
The general architecture of our IFRP framework is depicted in Figure~\ref{fig:pipeline}.

\vspace{0.05cm}
\noindent{\bf{Style Removal Network. }}
As the goal of face recovery is to generate a photorealistic destylized image, a generative network should be able to remove various styles of portraits without loosing the identity information. To this end, we propose the SRN block which employs a fully convolutional autoencoder (a downsampling encoder and an upsampling decoder) with skip connections and STN layers. Figure~\ref{fig:pipeline} shows the architecture of our SRN block (the blue frame).

The autoencoder learns a deterministic mapping to transform images from the space of portraits into some latent space (via an encoder), and a mapping from the latent space to the space of real faces (via a decoder). 
In this manner, the encoder extracts high-level features of  unaligned stylized faces and transforms them into a feature vectors of some latent real face domain while the decoder synthesizes photorealistic faces from these feature vectors.

\begin{figure*}[t]
\centering
\includegraphics[width=0.13\linewidth]{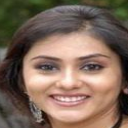}
\includegraphics[width=0.13\linewidth]{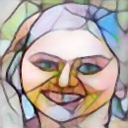}
\includegraphics[width=0.13\linewidth]{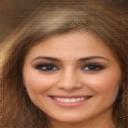}
\includegraphics[width=0.13\linewidth]{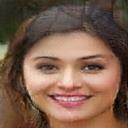}
\includegraphics[width=0.13\linewidth]{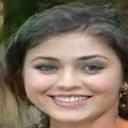}
\includegraphics[width=0.13\linewidth]{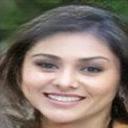}
\includegraphics[width=0.13\linewidth]{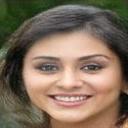}\\
\includegraphics[width=0.13\linewidth]{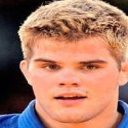}
\includegraphics[width=0.13\linewidth]{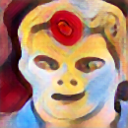}
\includegraphics[width=0.13\linewidth]{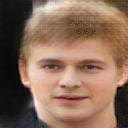}
\includegraphics[width=0.13\linewidth]{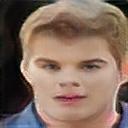}
\includegraphics[width=0.13\linewidth]{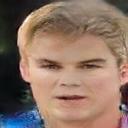}
\includegraphics[width=0.13\linewidth]{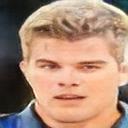}
\includegraphics[width=0.13\linewidth]{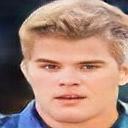}\\
\includegraphics[width=0.13\linewidth]{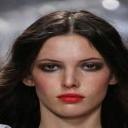}
\includegraphics[width=0.13\linewidth]{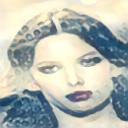}
\includegraphics[width=0.13\linewidth]{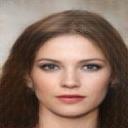}
\includegraphics[width=0.13\linewidth]{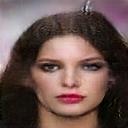}
\includegraphics[width=0.13\linewidth]{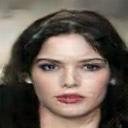}
\includegraphics[width=0.13\linewidth]{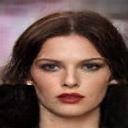}
\includegraphics[width=0.13\linewidth]{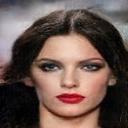}\\
\subfigure[]{\label{fig:ablationA}\scalebox{1}[1]{\includegraphics[width=0.13\linewidth]{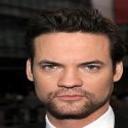}}}
\subfigure[]{\label{fig:ablationB}\scalebox{1}[1]{\includegraphics[width=0.13\linewidth]{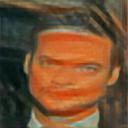}}}
\subfigure[]{\label{fig:ablationC}\scalebox{1}[1]{\includegraphics[width=0.13\linewidth]{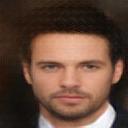}}}
\subfigure[]{\label{fig:ablationD}\scalebox{1}[1]{\includegraphics[width=0.13\linewidth]{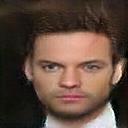}}}
\subfigure[]{\label{fig:ablationE}\scalebox{1}[1]{\includegraphics[width=0.13\linewidth]{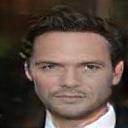}}}
\subfigure[]{\label{fig:ablationF}\scalebox{1}[1]{\includegraphics[width=0.13\linewidth]{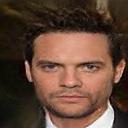}}}
\subfigure[]{\label{fig:ablationG}\scalebox{1}[1]{\includegraphics[width=0.13\linewidth]{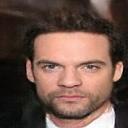}}}
\caption{The impact of various components of our network on the performance. (a) Ground truth face images. (b) Unaligned input portraits. (c) Results without the use of skip connections/residual blocks in the SRN block similar to the autoencoder in~\cite{Shiri2017FaceD}. (d) Results with the use of U-net autoencoder. The SRN block similar to the autoencoder in~\cite{ronneberger2015u} is used.  (e) Results with skip connections but without residual blocks in the SRN unit. 
(f) Results without STN layers in the SRN block. (g) Our final results with skip connections/residual blocks in the SRN.}
\label{fig:ablation}  
\end{figure*}

Moreover, we symmetrically link convolutional and deconvolutional layers via skip-layer connections~\cite{long2015fully}. These skip connections pass high-resolution visual details of portraits from convolutional to deconvolutional layers, leading to a good quality recovery.
In detail, each skip connection comprises three residual blocks. Due to the usage of residual blocks, our network can remove the styles of input portraits and increase the visual quality as shown in Figure~\ref{fig:ablationG}. In contrast, the same network but without skip connections tends to produce blurry/fuzzy face images as shown in Figure~\ref{fig:ablationC}. Figure~\ref{fig:ablation} shows that the visual quality improves as components of our architecture are enabled one-by-one.

As input stylized faces are often misaligned due to in-plane rotations, translations and scale changes, we incorporate Spatial Transformer Networks (STNs)~\cite{jaderberg2015spatial} (green blocks in Figure~\ref{fig:pipeline}) into the SRN. 
The STN layer can estimate the motion parameters of face images and warp them to the so-called canonical view. Thus, our method does not require the use of facial landmarks or 3D face models (often used for face alignment).  Figure \ref{fig:ablationG} shows that these intermediate STN layers help compensate for misalignment of the input portraits (however, their use is discretionary).
The architecture of our STN layers is given in the Appendix \ref{sec:a1}.

For appearance similarity between the recovered faces and their RF ground truth counterparts, we exploit a pixel-wise $\ell_2$ loss and an identity-preserving loss. The pixel-wise $\ell_2$ loss enforces intensity-based similarity between images of recovered faces and their ground truth images. The autoencoder supervised by the $\ell_2$ loss tends to produce oversmooth results as shown in Figure~\ref{fig:DisC}. 
For the identity-preserving loss, we use FaceNet~\cite{schroff2015facenet} to extract features from images (see Section~\ref{sec:training} for more details), and then we compare the Euclidean distance between feature maps of two images. In this way, we encourage the feature similarity between  recovered faces and their ground truth counterparts. 
Without the identity-preserving loss, the network produces random artifacts that resemble facial details, such as wrinkles, as shown in Figure~\ref{fig:DisD}.

\vspace{0.05cm}
\noindent{\bf{Discriminative Network. }}
Using only the pixel-wise distance between the recovered faces and their ground truth real counterparts leads to oversmooth results, as shown in Figure~\ref{fig:DisC}. To obtain appealing visual results, we introduce a discriminator, which forces recovered faces to reside in the same latent space as real faces. 
Our proposed DN is composed of convolutional layers and fully connected layers, as illustrated in Figure~\ref{fig:pipeline} (the red frame). 
The discriminative loss, also known as the adversarial loss, penalizes the discrepancy between the distributions of recovered and real faces. This loss is also used to update the parameters of the SRN block (we alternate over updates of the parameters of SRN and DN). Figure~\ref{fig:DisD} shows the impact of the adversarial loss on the final results.

\vspace{0.05cm}
\noindent{\bf{Identity Preservation. }}
With the adversarial loss, the SRN is able to generate high-frequency facial content. However, the results often lack details of identities such as the beard or wrinkles, as illustrated in Figure~\ref{fig:DisD}. 
A possible way to address this issue is to constrain the recovered face images and the ground truth face images to share the same face-related visual features \eg, FaceNet features~\cite{schroff2015facenet}.

\subsection{Training Details}
\label{sec:training}
To train our IFRP network in an end-to-end fashion, we require a large number of SF/RF training image pairs. For each RF, we synthesize different unaligned SF images according to chosen artistic styles to obtain SF/RF training pairs $(\mI_s,\mI_r)$. As described in Section~\ref{Sec:Data}, we only use stylized faces from three distinct styles in the training stage. 

Motivated by the ideas of Gatys \etal~\cite{gatys2016image} and Johnson \etal~\cite{johnson2016perceptual}, we construct so-called identity-preserving loss. Specifically, we compute the Euclidean distance between the feature maps of the recovered and ground truth images. These feature maps are obtained from the ReLU activations of FaceNet~\cite{schroff2015facenet}.

\begin{figure}[t]
\centering
\includegraphics[width=0.22\linewidth]{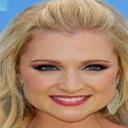}
\includegraphics[width=0.22\linewidth]{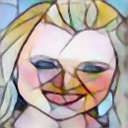}
\includegraphics[width=0.22\linewidth]{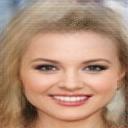}
\includegraphics[width=0.22\linewidth]{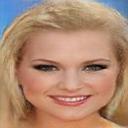}\\ \vspace{-0.4em}
\subfigure[]{\label{fig:VGGA}\scalebox{1}[1]{\includegraphics[width=0.22\linewidth]{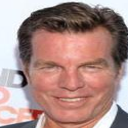}}}
\subfigure[]{\label{fig:VGGB}\scalebox{1}[1]{\includegraphics[width=0.22\linewidth]{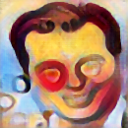}}}
\subfigure[]{\label{fig:VGGC}\scalebox{1}[1]{\includegraphics[width=0.22\linewidth]{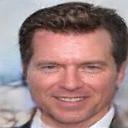}}}
\subfigure[]{\label{fig:VGGD}\scalebox{1}[1]{\includegraphics[width=0.22\linewidth]{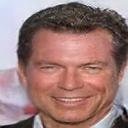}}}
\caption{The identity preservation loss: comparison of VGG-19 vs. FaceNet. (a) Ground truth. (b) Unaligned input portraits from the test dataset. (c) Recovered faces using VGG-19~\cite{simonyan2014very}. (d) Our final results using FaceNet~\cite{schroff2015facenet}.}
\label{fig:VGG}  
\end{figure}

Our previous work~\cite{Shiri2017FaceD} uses only the Euclidean loss to compare the generated and ground truth images which results in blurry images. In this work, we use the FaceNet network for the identity preservation loss and compare FaceNet to VGG-19 which is pre-trained on the large-scale ImageNet dataset containing objects. In contrast, FaceNet is pre-trained on a large dataset of 200 million face identities and 800 million pairs of face images. Therefore, FaceNet can capture visually meaningful facial features. As shown in Figure~\ref{fig:VGGD}, with the help of FaceNet, our results achieve higher fidelity and better consistency with respect to the ground truth face images. Figure~\ref{fig:VGGC} shows the results for VGG-19.

With FaceNet, we can preserve the identity information by encouraging the feature similarity between the generated and ground truth faces.  
We combine the pixel-wise loss, the adversarial loss and the identity-preserving loss together as our final loss function to train our network. Figure~\ref{fig:DisE} illustrates that, with the help of the identity-preserving loss, our IFRP network can recover satisfying identity-preserving images. 
Below we explain each loss individually.

\vspace{0.05cm}
\noindent{\bf{Pixel-wise Intensity Similarity Loss. }}
Our goal is to train our feed-forward SRN to produce an aligned photorealistic face image from any given stylized unaligned portrait. To achieve this, we force the recovered face image $\mIh_r$ to be similar to its ground truth counterpart $\mI_r$. We denote the output of our SRN as $\mG_\mTheta(\mI_s)$. Since the STN layers are interwoven with the layers of our autoencoder, we optimize the parameters of the autoencoder and the STN layers simultaneously. 
The pixel-wise loss function $\mathcal{L}_{\small{\rm pix}}$ between $\mIh_r$ and $\mI_r$ is expressed as:
\begin{equation}
\label{eqn:pix}
\begin{split}
\mathcal{L}_{\rm pix}(\mTheta)\!&=\!\mathbb{E}_{(\mI_s,\mI_r)\sim p(\mI_s,\mI_r)} \|\mG_{\mTheta}(\mI_s) - \mI_r\|_F^2,
\end{split}
\end{equation}
where $p(\mI_s,\mI_r)$ represents the joint distribution of the SF and RF images in the training dataset, and $\mTheta$ denotes the parameters of the SRN block.

\vspace{0.05cm}
\noindent{\bf{Identity-preserving Loss. }}
To obtain convincing identity-preserving results, we propose an identity-preserving loss to take the form of the Euclidean distance between the features of recovered face image $\mIh_r= \mG_{\mTheta}(\mI_s)$ and the ground truth face image $\mI_r$.  
The identity-preserving loss $\mathcal{L}_{id}$ is given as: 
\begin{equation}
\label{Identity}
\begin{split}
&\!\mathcal{L}_{\rm id}(\mTheta)=  \mathbb{E}_{(\mI_s,\mI_r)\sim p(\mI_s,\mI_r)}\|\vpsi(\mG_{\mTheta}(\mI_s))-\vpsi(\mI_r)\|^2_{F},\!\!
\end{split}\!\!\!
\end{equation}
where $\vpsi(\cdot)$ denotes the extracted feature maps from the layer ReLU3-2 of the FaceNet model with respect to some input image.

\vspace{0.05cm}
\noindent{\bf{Discriminative Loss. }}
Motivated by the idea of~\cite{Goodfellow2014,denton2015deep,radford2015unsupervised}, we aim to make the discriminative network $\mD_\mPhi$ fail to distinguish recovered face images from  ground truth face images.
Therefore, the parameters of the discriminator $\mPhi$ are updated by minimizing $\mathcal{L}_{\rm dis}$, expressed as:
\begin{equation}
\label{eqn:discr}
\begin{split}
\mathcal{L}_{\rm dis}(\mPhi)\!=&-\!\mathbb{E}_{\mI_r\sim p(\mI_r)}[\log \mD_\mathcal{\mPhi}(\mI_r)]\\
&-\! \mathbb{E}_{\mIh_r\sim p(\mIh_r)}[\log(1\!-\!\mD_\mathcal{\mPhi}(\mIh_r))],
\end{split}\!\!\!
\end{equation}
where $p(\mI_r)$ and $p(\mIh_r)$ indicate the distributions of real and recovered face images, respectively, and $\mD_\mPhi(\mI_r)$ and $\mD_\mPhi(\mIh_r)$ are the outputs of $\mD_\mPhi$ for real and recovered face images. The $\mathcal{L}_{\rm dis}$ loss is also backpropagated with respect to the parameters $\mTheta$ of the SRN block. 

Our SNR loss is a weighted sum of three terms: the pixel-wise loss, the adversarial loss, and the identity-preserving loss. The parameters $\mTheta$ are obtained by minimizing the final objective function of the SRN loss given below:
\begin{equation}
\begin{split}
\!\!\mathcal{L}_{\rm SNR}(\mTheta)=& \mathcal{L}_{\rm pix} \!+\!\lambda\ \mathcal{L}_{\rm dis} \!+\!\eta\ \mathcal{L}_{id} \\
\!=& \mathbb{E}_{(\mI_s,\mI_r)\sim p(\mI_s,\mI_r)} \|\mG_{\mTheta}(\mI_s) - \mI_r\|_F^2\\
\!+& \lambda\ \mathbb{E}_{\mI_s\sim p(\mI_s))}[\log\!\mD_\mPhi(\mG_{\mTheta}(\mI_s))]\\
\!+& \eta\ \mathbb{E}_{(\mI_s,\mI_r)\sim p(\mI_s,\mI_r)}\|\vpsi(\mG_{\mTheta}(\mI_s))-\vpsi(\mI_r)\|^2_{F},\!\!\!\!\!\!\!\!
\end{split}
\end{equation}
where $\lambda$ and $\eta$ are trade-off parameters for the discriminator and the identity-preserving losses, respectively, and $p(\mI_s)$ is the distribution of stylized face images. 

Since both $\mG_{\mTheta}(\cdot)$ and $\mD_{\mPhi}(\cdot)$ are differentiable functions, the error can be backpropagated w.r.t. $\mTheta$ and $\mPhi$ by the use of the Stochastic Gradient Descent (SGD) combined with the Root Mean Square Propagation (RMSprop)~\cite{Hinton}, which helps our algorithm converge faster. %

\subsection{Implementation Details}
The discriminative network $DN$ is only required in the training phase. In the testing phase, we take SP portraits as inputs and feed them to SRN. The outputs of SRN are the recovered photorealistic face images. We employ convolutional layers with kernels of size $4\times4$ and stride $2$ in the encoder and deconvolutional layers with kernels of size $4\times4$ and stride $2$ in the decoder. The feature maps in our encoder are passed to the decoder by skip connections. The batch normalization procedure is applied after our convolutional and deconvolutional layers except for the last deconvolutional layer, similar to the models described in ~\cite{Goodfellow2014,radford2015unsupervised}. For the non-linear activation function, we use the leaky rectifier with piecewise linear units (leakyReLU~\cite{maas2013rectifier}), for which the weight of negative slope is set to $0.2$.

Our network is trained with a mini-batch size of 64, the learning rate set to $10^{-3}$ and the decay rate set to $10^{-2}$.
In all our experiments, parameters $\lambda$ and $\eta$ are set to $10^{-2}$ and $10^{-3}$. As the iterations progress, the images of output faces will be more similar to the ground truth. Hence, we gradually reduce the effect of the discriminative network by decreasing $\lambda$. Thus,
$\lambda^{n} = \max\{\lambda\cdot 0.995^n, \lambda/2 \}$,
where $n$ is the epoch index. 
The strategy in which we decrease $\lambda$ not only enriches the impact of the pixel-level similarity but also helps preserve the discriminative information in the SRN during training.
We also decrease $\eta$ to reduce the impact of the identity-preserving constraint after each iteration. Thus, $\eta^{n} = \max\{\eta\cdot 0.995^n, \eta/2\}$.

\begin{figure*}
\centering
\subfigure[]{\label{fig:dataset1}\scalebox{1}[1]{\includegraphics[width=0.095\linewidth]{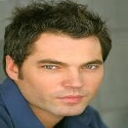}}}\\\vspace{-0.8em}
\subfigure[]{\label{fig:dataseta}\scalebox{1}[1]{\includegraphics[width=0.095\linewidth]{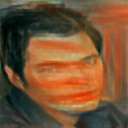}}}
\subfigure[]{\label{fig:datasetb}\scalebox{1}[1]{\includegraphics[width=0.095\linewidth]{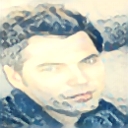}}}
\subfigure[]{\label{fig:datasetc}\scalebox{1}[1]{\includegraphics[width=0.095\linewidth]{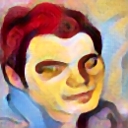}}}
\subfigure[]{\label{fig:datasetd}\scalebox{1}[1]{\includegraphics[width=0.095\linewidth]{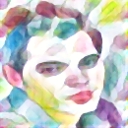}}}
\subfigure[]{\label{fig:datasetdd}\scalebox{1}[1]{\includegraphics[width=0.095\linewidth]{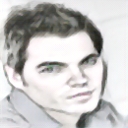}}}
\subfigure[]{\label{fig:datasete}\scalebox{1}[1]{\includegraphics[width=0.095\linewidth]{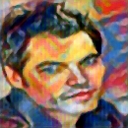}}}
\subfigure[]{\label{fig:datasetf}\scalebox{1}[1]{\includegraphics[width=0.095\linewidth]{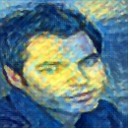}}}
\subfigure[]{\label{fig:datasetg}\scalebox{1}[1]{\includegraphics[width=0.095\linewidth]{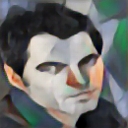}}}
\subfigure[]{\label{fig:dataseth}\scalebox{1}[1]{\includegraphics[width=0.095\linewidth]{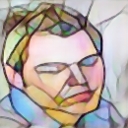}}} 
\subfigure[]{\label{fig:datasethh}\scalebox{1}[1]{\includegraphics[width=0.095\linewidth]{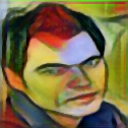}}} \vspace{-0.5em}\vspace{-0.5em}
\caption{Samples from the synthesized dataset. (a) The ground truth aligned real face image. (b)-(k) The synthesized unaligned portraits form~\emph{Scream, Wave, Candy, Feathers, Sketch, Composition VII,Starry night, Udnie, Mosaic } and \emph{la Muse} styles which have been used for training and testing our network.}
\label{fig:dataset}
\end{figure*}

As our method is of feed-forward nature (no optimization is required at the test time), it takes 8 ms to destylize a 128$\times$128 image. 

\section{Synthesized Dataset and Preprocessing}
\label{Sec:Data}
To train our IFRP network and avoid overfitting, a large number of SF/RF image pairs are required.  
To generate a dataset of such pairs, similar to~\cite{Shiri2017FaceD}, we use the Celebrity dataset (CelebA)~\cite{Liu2015faceattributes}. 
Firstly, we randomly select 110K faces from the CelebA dataset for training and 2K face images for testing. The original size of images is $178\!\times\!218$ pixels. Subsequently, we crop/extract the center of each image and resize it to $128\!\times\!128$ pixels. We use such cropped images as our RF ground truth face images $\mI_r$. Lastly, we apply affine transformations to the aligned ground truth face images to generate in-plane unaligned face images.  

Moreover, to synthesize our training dataset, we retrain the real-time style transfer network \cite{johnson2016perceptual} for different artworks. We use only three distinct styles, \emph{Scream, Candy} and \emph{Mosaic} for synthesizing our training dataset. The procedure detailing how we selected these styles is explained in Section~\ref{metric}. 
We also use 2K unaligned ground truth face images to synthesize 20K SF images from ten diverse styles (\emph{Scream, Wave, Candy, Feathers, Sketch, Composition VII, Starry night, Udnie, Mosaic} and \emph{la Muse}) as our testing dataset. Note that we also include artistic sketches as an unseen style into our test dataset. Some stylized face images used for training and testing are shown in Figure~\ref{fig:dataset}. Lastly, we emphasize that there is no overlap between the training and testing datasets.

\begin{table}\renewcommand{\arraystretch}{1}
\centering
\caption{The number of training styles and the corresponding training times.}
\begin{tabular}{>{\centering\arraybackslash}m{0.218\linewidth}|>{\centering\arraybackslash}m{0.194\linewidth}|>{\centering\arraybackslash}m{0.06\linewidth}|>{\centering\arraybackslash}m{0.06\linewidth}|>{\centering\arraybackslash}m{0.06\linewidth}|>{\centering\arraybackslash}m{0.02\linewidth}}
\hline
\multirow{2}{*}{\pbox{3cm}{Number of\\Training Styles}} & \multirow{2}{*}{\pbox{3cm}{Training time\\per epoch}} & \multicolumn{2}{c|}{Seen Styles} & \multicolumn{2}{c}{Unseen Styles}\\
\cline{3-6}
\vspace{1mm}
\centering
 &  & SSIM &FSIM &SSIM &FSIM\\
\hline
1 Style & 1:49' & 0.69 & 0.72 & 0.54 & 0.66\\
2 Styles & 3:54' & 0.70& 0.77 & 0.60 & 0.78\\
3 Styles & 5:20' &  0.72 &  0.88  &  0.68 & 0.84\\
4 Styles & 7:05'  & 0.72 & 0.88 & 0.68 & 0.85\\
5 Styles & 9:47'  & 0.73 & 0.88 & 0.69 & 0.85\\
\hline
\end{tabular}
\label{tab15}
\end{table}
 
\begin{table}\renewcommand{\arraystretch}{1}
\centering
\caption{The number of training epochs vs. the number of styles (for the same number of augmentations).}
\begin{tabular}{>{\centering\arraybackslash}m{0.30\linewidth}|>{\centering\arraybackslash}m{0.3\linewidth}|>{\centering\arraybackslash}m{0.2\linewidth}}
\hline
\multirow{2}{*}{\pbox{3cm}{Number of\\Training Styles}} & \multicolumn{2}{c}{Number of epochs} \\
\cline{2-3}
\vspace{1mm}
\centering
 & Without STNs & With STN \\
\hline
1 Style & 203 & 180 \\
2 Styles & 181 & 159\\
3 Styles & 165 & 143\\
4 Styles & 150 & 135\\
5 Styles & 139 & 121\\
\hline
\end{tabular}
\label{tab150}
\end{table}

\section{Experiments}
\label{expriment}
Below we compare the performance of our approach qualitatively and quantitatively to the state-of-the-art methods. 
To conduct a fair comparison, we retrain  approaches~\cite{gatys2016image,johnson2016perceptual,li2016precomputed,isola2016image,zhu2017unpaired,Shiri2017FaceD} on our training dataset for the task of photorealistic face recovery from portraits.

\subsection{Style-Distance Metric}
\label{metric}
Generating/training on large numbers of styles  is impractical. Thus, we propose a style-distance metric to select the most difficult styles for the process of face recovery.
For this purpose,  we compute Gram matrices for various styles from feature maps of pre-trained VGG model~\cite{simonyan2014very}. Then, we measure the similarity of styles based on the Log-Euclidean distance~\cite{jayasumana2013kernel} between Gram matrices of style images and the average Gram matrix of all ground truth face images in our training dataset. 
As these Gram matrices reflect the style differences between input images~\cite{gatys2016image}, we choose three styles with the largest distances from the average Gram matrix of ground truth face images. According to the above criterion, we select \emph{Candy, Mosaic} and \emph{Scream} styles for training. 

Utilizing additional training styles can improve the quality of recovered images especially from unseen styles at a cost of extra training time. We have observed that using three training styles is optimal (using more styles does not improve the accuracy significantly). 
Table~\ref{tab15} summarizes the average SSIM and FSIM scores on the test dataset given different number of training styles. As the number of training styles increases, our network learns a better mapping between different genres of stylized portraits and ground truth face images. 
In order to help the network learn a mapping between unaligned and aligned data, we use STN layers that reduce the number of epochs needed for convergence. Table~\ref{tab150} shows that for training with 3 styles, our network converges after 165 epochs (without STNs) and 143 epochs (with STNs). We note that when the network is trained without STN layers, its visual performance is somewhat worse to the results which rely on STNs.

\begin{figure*}[!ht]
\begin{minipage}{0.089\linewidth}
\centering
\subfigure[RF]{\label{fig:cmp1rf}\scalebox{1}[1]
{\includegraphics[width=1.18\linewidth]{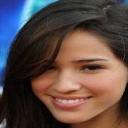}}}
\end{minipage}
\begin{minipage}{0.89\linewidth}
\centering
\hspace{0.5em}\includegraphics[width=0.118\linewidth]{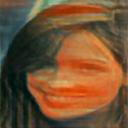}
\includegraphics[width=0.118\linewidth]{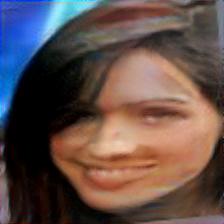}
\includegraphics[width=0.118\linewidth]{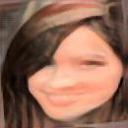}
\includegraphics[width=0.118\linewidth]{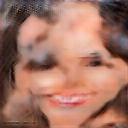}
\includegraphics[width=0.118\linewidth]{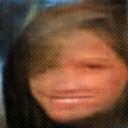}
\includegraphics[width=0.118\linewidth]{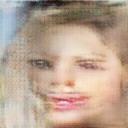}
\includegraphics[width=0.118\linewidth]{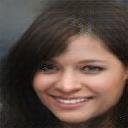}
\includegraphics[width=0.118\linewidth]{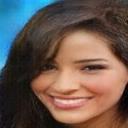}\\\vspace{0.1em}
\hspace{0.5em}\includegraphics[width=0.118\linewidth]{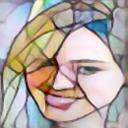}
\includegraphics[width=0.118\linewidth]{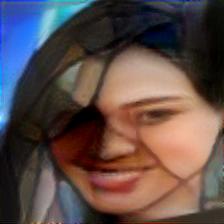}
\includegraphics[width=0.118\linewidth]{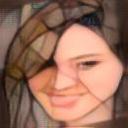}
\includegraphics[width=0.118\linewidth]{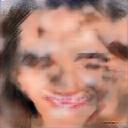}
\includegraphics[width=0.118\linewidth]{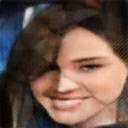}
\includegraphics[width=0.118\linewidth]{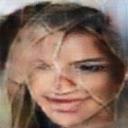}
\includegraphics[width=0.118\linewidth]{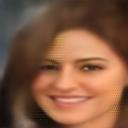}
\includegraphics[width=0.118\linewidth]{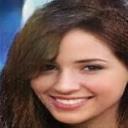}\\\vspace{0.1em}
\hspace{0.5em}\includegraphics[width=0.118\linewidth]{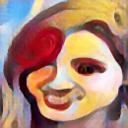}
\includegraphics[width=0.118\linewidth]{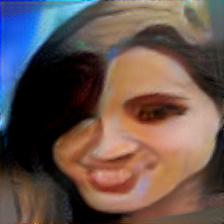}
\includegraphics[width=0.118\linewidth]{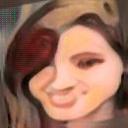}
\includegraphics[width=0.118\linewidth]{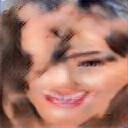}
\includegraphics[width=0.118\linewidth]{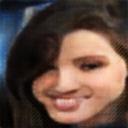}
\includegraphics[width=0.118\linewidth]{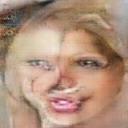}
\includegraphics[width=0.118\linewidth]{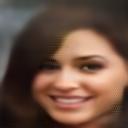}
\includegraphics[width=0.118\linewidth]{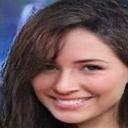}\\\vspace{0.1em}
\hspace{0.5em}\includegraphics[width=0.118\linewidth]{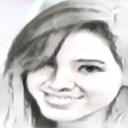}
\includegraphics[width=0.118\linewidth]{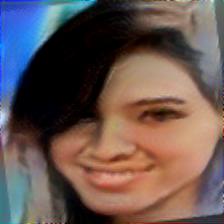}
\includegraphics[width=0.118\linewidth]{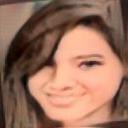}
\includegraphics[width=0.118\linewidth]{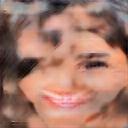}
\includegraphics[width=0.118\linewidth]{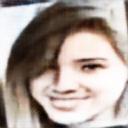}
\includegraphics[width=0.118\linewidth]{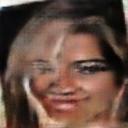}
\includegraphics[width=0.118\linewidth]{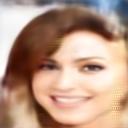}
\includegraphics[width=0.118\linewidth]{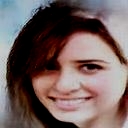}\\\vspace{0.1em}
\hspace{0.5em}\includegraphics[width=0.118\linewidth]{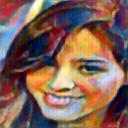}
\includegraphics[width=0.118\linewidth]{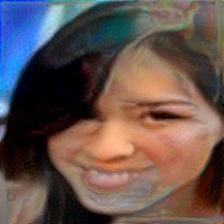}
\includegraphics[width=0.118\linewidth]{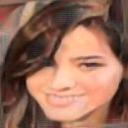}
\includegraphics[width=0.118\linewidth]{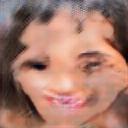}
\includegraphics[width=0.118\linewidth]{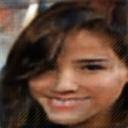}
\includegraphics[width=0.118\linewidth]{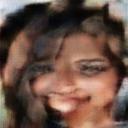}
\includegraphics[width=0.118\linewidth]{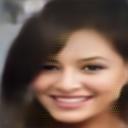}
\includegraphics[width=0.118\linewidth]{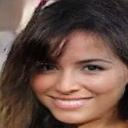}\\\vspace{0.1em}
\hspace{0.5em}\includegraphics[width=0.118\linewidth]{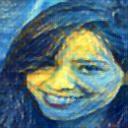}
\includegraphics[width=0.118\linewidth]{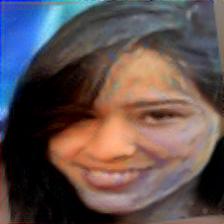}
\includegraphics[width=0.118\linewidth]{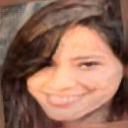}
\includegraphics[width=0.118\linewidth]{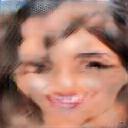}
\includegraphics[width=0.118\linewidth]{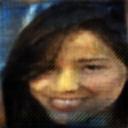}
\includegraphics[width=0.118\linewidth]{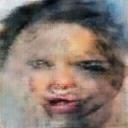}
\includegraphics[width=0.118\linewidth]{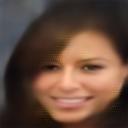}
\includegraphics[width=0.118\linewidth]{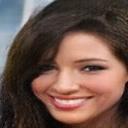}\\\vspace{0.1em}
\hspace{0.5em}\includegraphics[width=0.118\linewidth]{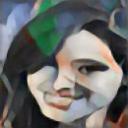}
\includegraphics[width=0.118\linewidth]{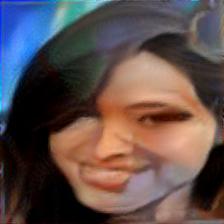}
\includegraphics[width=0.118\linewidth]{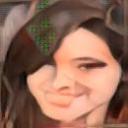}
\includegraphics[width=0.118\linewidth]{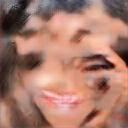}
\includegraphics[width=0.118\linewidth]{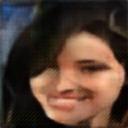}
\includegraphics[width=0.118\linewidth]{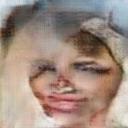}
\includegraphics[width=0.118\linewidth]{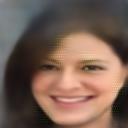}
\includegraphics[width=0.118\linewidth]{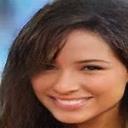}\\\vspace{-1.3mm}
\hspace{0.5em}\subfigure[SF]{\label{fig:cmp1b}\scalebox{1}[1]{\includegraphics[width=0.118\linewidth]{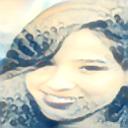}}}
\subfigure[\cite{gatys2016image}]{\label{fig:cmp1c}\scalebox{1}[1]
{\includegraphics[width=0.118\linewidth]{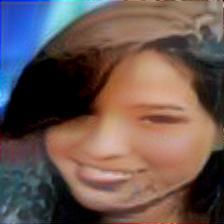}}}
\subfigure[\cite{johnson2016perceptual}]{\label{fig:cmp1d}\scalebox{1}[1]
{\includegraphics[width=0.118\linewidth]{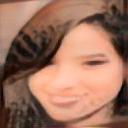}}}
\subfigure[\cite{li2016precomputed}]{\label{fig:cmp1e}\scalebox{1}[1]
{\includegraphics[width=0.118\linewidth]{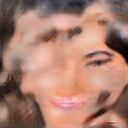}}}
\subfigure[\cite{isola2016image}]{\label{fig:cmp1f}\scalebox{1}[1]
{\includegraphics[width=0.118\linewidth]{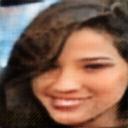}}}
\subfigure[\cite{zhu2017unpaired}]{\label{fig:cmp1g}\scalebox{1}[1]
{\includegraphics[width=0.118\linewidth]{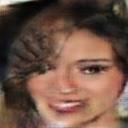}}}
\subfigure[\cite{Shiri2017FaceD}]{\label{fig:cmp1h}\scalebox{1}[1]
{\includegraphics[width=0.118\linewidth]{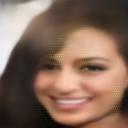}}}
\subfigure[Ours]{\label{fig:cmp1i}\scalebox{1}[1]
{\includegraphics[width=0.118\linewidth]{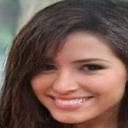}}}
\end{minipage}
\hfill
\vspace{-0.1cm}
\caption{Qualitative comparisons of the state-of-the-art methods. (a) The ground truth  face image. (b) Input portraits (from the test dataset) including the seen styles \emph{Scream, Mosaic} and \emph{Candy} as well as the unseen styles \emph{Sketch, Composition VII, Feathers, Udnie} and \emph{La Muse}. (c) Gatys \emph{et al.}'s method~\cite{gatys2016image}. (d) Johnson \emph{et al.}'s method~\cite{johnson2016perceptual}. (e) Li and Wand's method~\cite{li2016precomputed} (MGAN). (f) Isola \emph{et al.}'s method~\cite{isola2016image} (pix2pix). (g) Zhu \emph{et al.}'s method~\cite{zhu2017unpaired} (CycleGAN). (h) Shiri \emph{et al.}'s method~\cite{Shiri2017FaceD} (i) Our method.}
\label{fig:cmp1}
\end{figure*}

\subsection{Qualitative Evaluation}
We visually compare our approach against six methods detailed below. To help these methods achieve their best performance, we align SF images from the test dataset via a simple STN-based network prior to testing.

\begin{figure*}[!ht]
\hspace{2em}
\begin{minipage}{0.089\linewidth}
\centering
\subfigure[RF]{\label{fig:cmp2rf}\scalebox{1}[1]
{\includegraphics[width=1.18\linewidth]{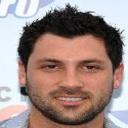}}} \hspace{2em}
\end{minipage}
\begin{minipage}{0.89\linewidth}
\centering
\hspace{0.5em}\includegraphics[width=0.118\linewidth]{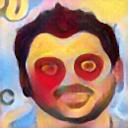}
\includegraphics[width=0.118\linewidth]{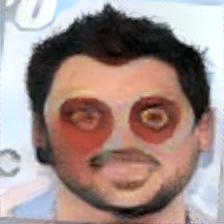}
\includegraphics[width=0.118\linewidth]{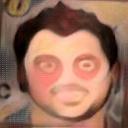}
\includegraphics[width=0.118\linewidth]{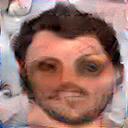}
\includegraphics[width=0.118\linewidth]{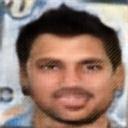}
\includegraphics[width=0.118\linewidth]{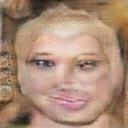}
\includegraphics[width=0.118\linewidth]{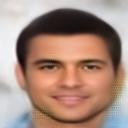}
\includegraphics[width=0.118\linewidth]{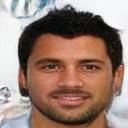}\\\vspace{0.1em}
\hspace{0.5em}\includegraphics[width=0.118\linewidth]{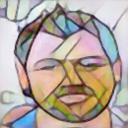}
\includegraphics[width=0.118\linewidth]{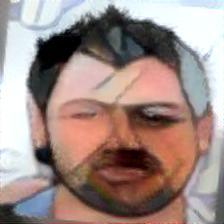}
\includegraphics[width=0.118\linewidth]{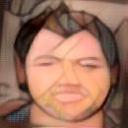}
\includegraphics[width=0.118\linewidth]{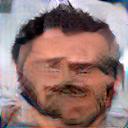}
\includegraphics[width=0.118\linewidth]{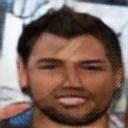}
\includegraphics[width=0.118\linewidth]{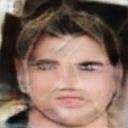}
\includegraphics[width=0.118\linewidth]{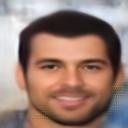}
\includegraphics[width=0.118\linewidth]{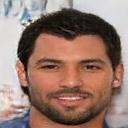}\\\vspace{0.1em}
\hspace{0.5em}\includegraphics[width=0.118\linewidth]{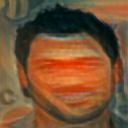}
\includegraphics[width=0.118\linewidth]{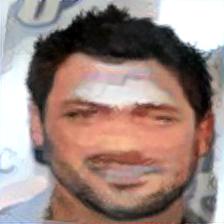}
\includegraphics[width=0.118\linewidth]{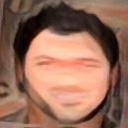}
\includegraphics[width=0.118\linewidth]{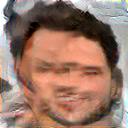}
\includegraphics[width=0.118\linewidth]{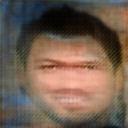}
\includegraphics[width=0.118\linewidth]{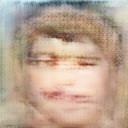}
\includegraphics[width=0.118\linewidth]{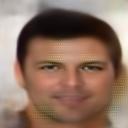}
\includegraphics[width=0.118\linewidth]{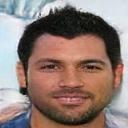}\\\vspace{0.1em}
\hspace{0.5em}\includegraphics[width=0.118\linewidth]{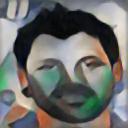}
\includegraphics[width=0.118\linewidth]{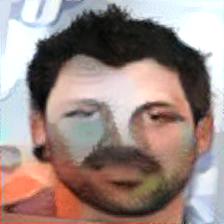}
\includegraphics[width=0.118\linewidth]{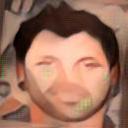}
\includegraphics[width=0.118\linewidth]{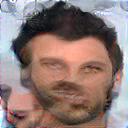}
\includegraphics[width=0.118\linewidth]{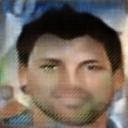}
\includegraphics[width=0.118\linewidth]{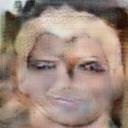}
\includegraphics[width=0.118\linewidth]{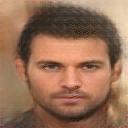}
\includegraphics[width=0.118\linewidth]{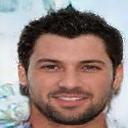}\\\vspace{0.1em}
\hspace{0.5em}\includegraphics[width=0.118\linewidth]{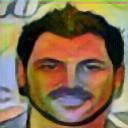}
\includegraphics[width=0.118\linewidth]{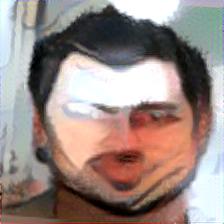}
\includegraphics[width=0.118\linewidth]{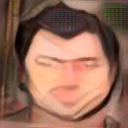}
\includegraphics[width=0.118\linewidth]{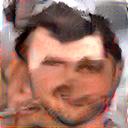}
\includegraphics[width=0.118\linewidth]{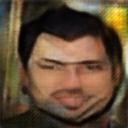}
\includegraphics[width=0.118\linewidth]{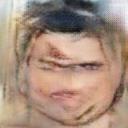}
\includegraphics[width=0.118\linewidth]{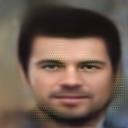}
\includegraphics[width=0.118\linewidth]{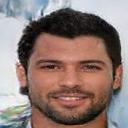}\\\vspace{0.1em}
\hspace{0.5em}\includegraphics[width=0.118\linewidth]{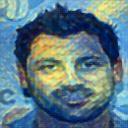}
\includegraphics[width=0.118\linewidth]{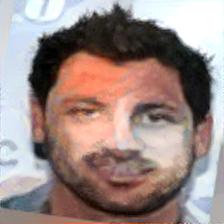}
\includegraphics[width=0.118\linewidth]{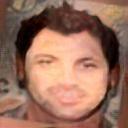}
\includegraphics[width=0.118\linewidth]{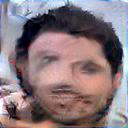}
\includegraphics[width=0.118\linewidth]{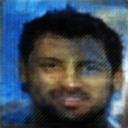}
\includegraphics[width=0.118\linewidth]{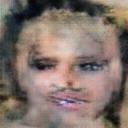}
\includegraphics[width=0.118\linewidth]{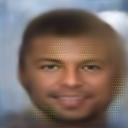}
\includegraphics[width=0.118\linewidth]{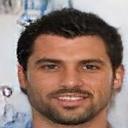}\\\vspace{0.1em}
\hspace{0.5em}\includegraphics[width=0.118\linewidth]{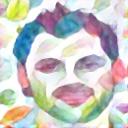}
\includegraphics[width=0.118\linewidth]{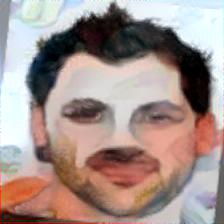}
\includegraphics[width=0.118\linewidth]{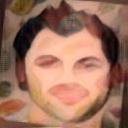}
\includegraphics[width=0.118\linewidth]{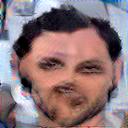}
\includegraphics[width=0.118\linewidth]{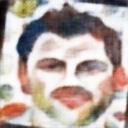}
\includegraphics[width=0.118\linewidth]{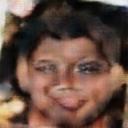}
\includegraphics[width=0.118\linewidth]{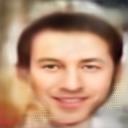}
\includegraphics[width=0.118\linewidth]{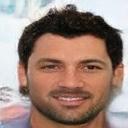}\\\vspace{-1.3mm}
\hspace{0.5em}\subfigure[SF]{\label{fig:cmp2b}\scalebox{1}[1]{\includegraphics[width=0.118\linewidth]{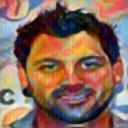}}}
\subfigure[\cite{gatys2016image}]{\label{fig:cmp2c}\scalebox{1}[1]
{\includegraphics[width=0.118\linewidth]{figs/200162_Com.jpg}}}
\subfigure[\cite{johnson2016perceptual}]{\label{fig:cmp2d}\scalebox{1}[1]
{\includegraphics[width=0.118\linewidth]{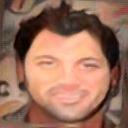}}}
\subfigure[\cite{li2016precomputed}]{\label{fig:cmp2e}\scalebox{1}[1]
{\includegraphics[width=0.118\linewidth]{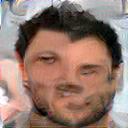}}}
\subfigure[\cite{isola2016image}]{\label{fig:cmp2f}\scalebox{1}[1]
{\includegraphics[width=0.118\linewidth]{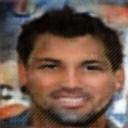}}}
\subfigure[\cite{zhu2017unpaired}]{\label{fig:cmp2g}\scalebox{1}[1]
{\includegraphics[width=0.118\linewidth]{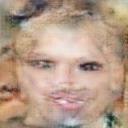}}}
\subfigure[\cite{Shiri2017FaceD}]{\label{fig:cmp2h}\scalebox{1}[1]
{\includegraphics[width=0.118\linewidth]{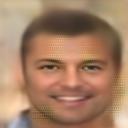}}}
\subfigure[Ours]{\label{fig:cmp2i}\scalebox{1}[1]
{\includegraphics[width=0.118\linewidth]{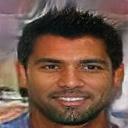}}}
\end{minipage}
\hfill
\vspace{-0.1cm}
\caption{Qualitative comparisons of the state-of-the-art methods. (a) The ground truth real face image. (b) Input portraits (from the test dataset) including the seen styles \emph{Candy, Mosaic} and \emph{Scream} as well as the unseen styles \emph{Udnie, La Muse, Starry, Feathers} and \emph{Composition VII}. (c) Gatys \emph{et al.}'s method~\cite{gatys2016image}. (d) Johnson \emph{et al.}'s method~\cite{johnson2016perceptual}. (e) Li and Wand's method~\cite{li2016precomputed} (MGAN). (f) Isola \emph{et al.}'s method~\cite{isola2016image} (pix2pix). (g) Zhu \emph{et al.}'s method~\cite{zhu2017unpaired} (CycleGAN). (h) Shiri \emph{et al.}'s method~\cite{Shiri2017FaceD} (i) Our method.}
\label{fig:cmp2}
\end{figure*}

Gatys \emph{et al.}~\cite{gatys2016image} is an image optimization-based style transfer method which does not have any training stage.
This method captures the correlation between feature maps of the portrait and the synthesized face via Gram matrices constructed from features extracted across several layers of a CNN pipeline. Thus, spatial structure of face images cannot be preserved by this approach. 
As shown in Figures~\ref{fig:cmp1c} and \ref{fig:cmp2c}, the network fails to remove various aspects of artistic styles and thus produces visually unconvincing results.

We also retrain the approach of Johnson \emph{et al.} \cite{johnson2016perceptual} for destylization. Due to the use of correlation statistics captured via the Gram matrix, their network also generates distorted facial details and produces unnatural artifacts. As Figures~\ref{fig:cmp1d} and \ref{fig:cmp2d} show, the facial details are blurred and the skin colors are not homogeneous. Moreover, Figure~\ref{fig:cmp2d} shows many images containing unnaturally looking eyes due to poor destylization abilities of approach  \cite{johnson2016perceptual}.

MGAN~\cite{li2016precomputed} is a patch-based style transfer method. We retrain this network for the purpose of the face recovery. As this method is trained on RF/SF patches, it cannot capture the global appearance of faces. As shown in Figures~\ref{fig:cmp1e} and \ref{fig:cmp2e}, this method produces distorted results and the facial colors are inconsistent. In contrast, our method successfully captures the global appearance of faces and generates consistent facial colors.

Isola~\emph{et al.}~\cite{isola2016image} train a ``U-net'' generator augmented with a PatchGAN discriminator in an adversarial framework, known as ``pix2pix''. Since the patch-based discriminator is trained to classify whether an image patch is sampled from the distr. of real faces or not, this network does not take the global appearance of faces into account.
In addition, U-net concatenates low-level features from the bottom layers of the encoder with the features in the decoder to generate face images. As the low-level features of input images are passed to the output, U-net fails to eliminate  artistic styles in face images.
As shown in Figures~\ref{fig:cmp1f} and \ref{fig:cmp2f}, pix2pix can generate acceptable results for the seen styles but fails to remove the unseen styles and thus it produces obvious artifacts. 


CycleGAN~\cite{zhu2017unpaired} is an image-to-image translation method that uses unpaired datasets. This network provides a mapping between two different domains by the use of a cycle-consistency loss.
Since CycleGAN also employs a patch-based discriminator, this network cannot capture the global appearance of faces. As CycleGAN employs unpaired face datasets for RF and SF images, the low-level features of the stylized  and recovered faces are uncorrelated.
Thus, CycleGAN is not suitable for transferring stylized portraits to photorealistic images. As shown in Figures~\ref{fig:cmp1g} and ~\ref{fig:cmp2g}, this method produces distorted face images and it does not preserve the identities of faces in the input images.

Our first destylization approach \cite{Shiri2017FaceD} does not exploit an identity-preserving loss as it employs only a simple autoencoder to recover  photorealistic face images. In contrast, in this paper we study an identity-preserving loss that helps recover  photorealistic face images which preserve  underlying identities. 
We utilize 330K pairs of SF/RF face images. Our IFRP method is  robust in terms of recovery of realistic faces. As shown in Figure~\ref{fig:opene}, our old method suffers for instance from poor recovery of hair color. As shown in the fourth row of Figures \ref{fig:cmp1c}--\ref{fig:cmp1h}, all methods, except for ours in Figure \ref{fig:cmp1i}, fail to recover the correct facial complexion. As shown in the fourth row of Figures~\ref{fig:cmp2e}--\ref{fig:cmp2h}, these methods cannot recover male's beard. In contrast, in Figure~\ref{fig:cmp2i}, our method is shown to recover well such an important facial feature.

Compared to our previous approach and other methods, our new method attains a higher fidelity and better consistency with regards to facial expressions and skin tones. Our network can preserve the identity of a subject given either seen or unseen styles, as shown in Figures~\ref{fig:cmp1i} and ~\ref{fig:cmp2i}.

\begin{figure*}[!ht]
\centering
\includegraphics[width=0.118\linewidth]{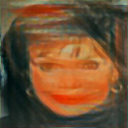}
\includegraphics[width=0.118\linewidth]{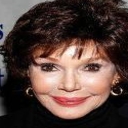}
\includegraphics[width=0.118\linewidth]{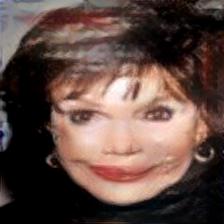}
\includegraphics[width=0.118\linewidth]{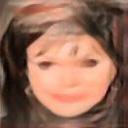}
\includegraphics[width=0.118\linewidth]{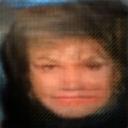}
\includegraphics[width=0.118\linewidth]{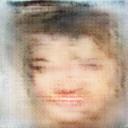}
\includegraphics[width=0.118\linewidth]{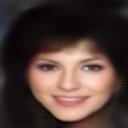}
\includegraphics[width=0.118\linewidth]{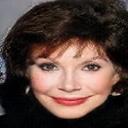}\\\vspace{0.1em}
\includegraphics[width=0.118\linewidth]{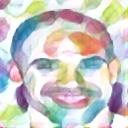}
\includegraphics[width=0.118\linewidth]{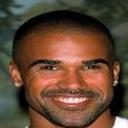}
\includegraphics[width=0.118\linewidth]{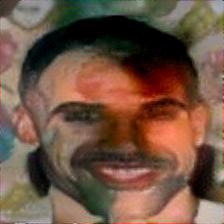}
\includegraphics[width=0.118\linewidth]{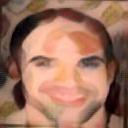}
\includegraphics[width=0.118\linewidth]{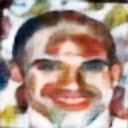}
\includegraphics[width=0.118\linewidth]{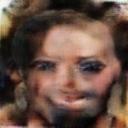}
\includegraphics[width=0.118\linewidth]{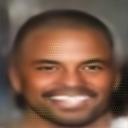}
\includegraphics[width=0.118\linewidth]{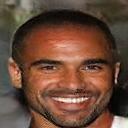}\\\vspace{0.1em}
\includegraphics[width=0.118\linewidth]{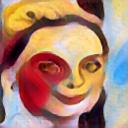}
\includegraphics[width=0.118\linewidth]{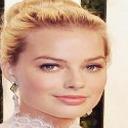}
\includegraphics[width=0.118\linewidth]{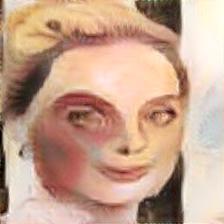}
\includegraphics[width=0.118\linewidth]{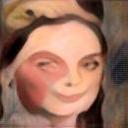}
\includegraphics[width=0.118\linewidth]{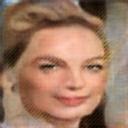}
\includegraphics[width=0.118\linewidth]{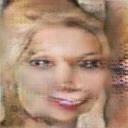}
\includegraphics[width=0.118\linewidth]{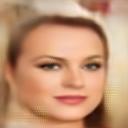}
\includegraphics[width=0.118\linewidth]{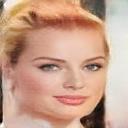}\\\vspace{0.1em}
\includegraphics[width=0.118\linewidth]{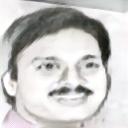}
\includegraphics[width=0.118\linewidth]{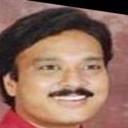}
\includegraphics[width=0.118\linewidth]{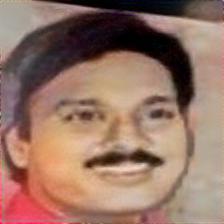}
\includegraphics[width=0.118\linewidth]{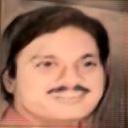}
\includegraphics[width=0.118\linewidth]{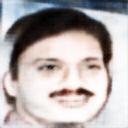}
\includegraphics[width=0.118\linewidth]{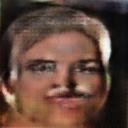}
\includegraphics[width=0.118\linewidth]{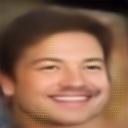}
\includegraphics[width=0.118\linewidth]{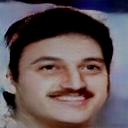}\\\vspace{0.1em}
\includegraphics[width=0.118\linewidth]{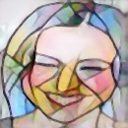}
\includegraphics[width=0.118\linewidth]{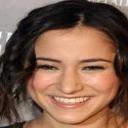}
\includegraphics[width=0.118\linewidth]{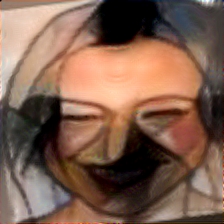}
\includegraphics[width=0.118\linewidth]{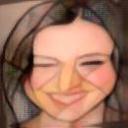}
\includegraphics[width=0.118\linewidth]{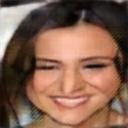}
\includegraphics[width=0.118\linewidth]{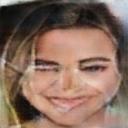}
\includegraphics[width=0.118\linewidth]{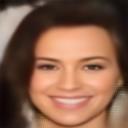}
\includegraphics[width=0.118\linewidth]{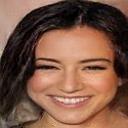}\\\vspace{0.1em}
\includegraphics[width=0.118\linewidth]{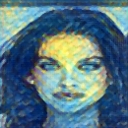}
\includegraphics[width=0.118\linewidth]{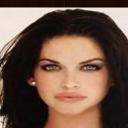}
\includegraphics[width=0.118\linewidth]{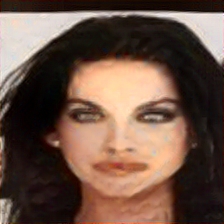}
\includegraphics[width=0.118\linewidth]{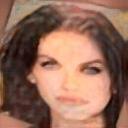}
\includegraphics[width=0.118\linewidth]{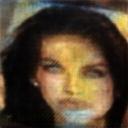}
\includegraphics[width=0.118\linewidth]{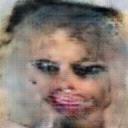}
\includegraphics[width=0.118\linewidth]{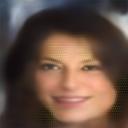}
\includegraphics[width=0.118\linewidth]{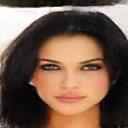}\\\vspace{0.1em}
\includegraphics[width=0.118\linewidth]{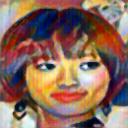}
\includegraphics[width=0.118\linewidth]{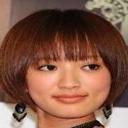}
\includegraphics[width=0.118\linewidth]{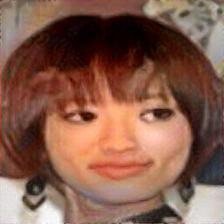}
\includegraphics[width=0.118\linewidth]{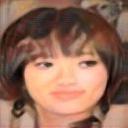}
\includegraphics[width=0.118\linewidth]{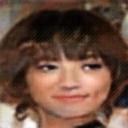}
\includegraphics[width=0.118\linewidth]{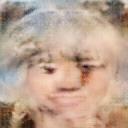}
\includegraphics[width=0.118\linewidth]{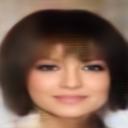}
\includegraphics[width=0.118\linewidth]{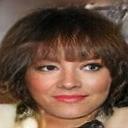}\\\vspace{0.1em}
\includegraphics[width=0.118\linewidth]{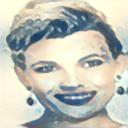}
\includegraphics[width=0.118\linewidth]{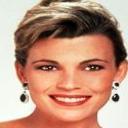}
\includegraphics[width=0.118\linewidth]{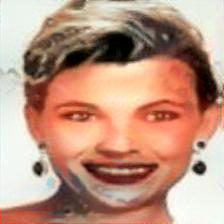}
\includegraphics[width=0.118\linewidth]{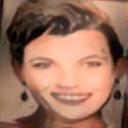}
\includegraphics[width=0.118\linewidth]{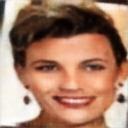}
\includegraphics[width=0.118\linewidth]{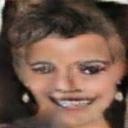}
\includegraphics[width=0.118\linewidth]{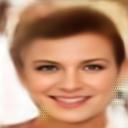}
\includegraphics[width=0.118\linewidth]{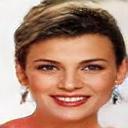}\\\vspace{0.1em}
\includegraphics[width=0.118\linewidth]{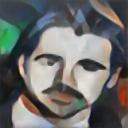}
\includegraphics[width=0.118\linewidth]{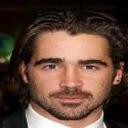}
\includegraphics[width=0.118\linewidth]{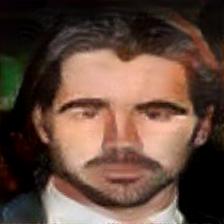}
\includegraphics[width=0.118\linewidth]{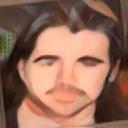}
\includegraphics[width=0.118\linewidth]{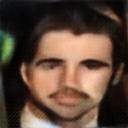}
\includegraphics[width=0.118\linewidth]{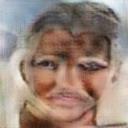}
\includegraphics[width=0.118\linewidth]{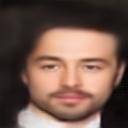}
\includegraphics[width=0.118\linewidth]{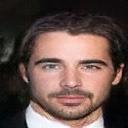}\\\vspace{-1.3mm}
\hspace{0.2cm}\subfigure[SF]{\label{fig:cmp6a}\scalebox{1}[1]{\includegraphics[width=0.118\linewidth]{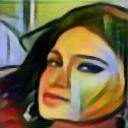}}}
\subfigure[RF]{\label{fig:cmp6b}\scalebox{1}[1]{\includegraphics[width=0.118\linewidth]{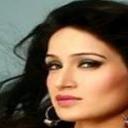}}}
\subfigure[\cite{gatys2016image}]{\label{fig:cmp6c}\scalebox{1}[1]
{\includegraphics[width=0.118\linewidth]{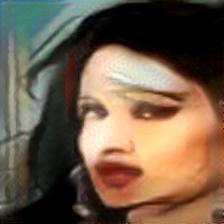}}}
\subfigure[\cite{johnson2016perceptual}]{\label{fig:cmp6d}\scalebox{1}[1]
{\includegraphics[width=0.118\linewidth]{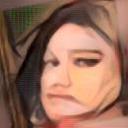}}}
\subfigure[\cite{isola2016image}]{\label{fig:cmp6e}\scalebox{1}[1]
{\includegraphics[width=0.118\linewidth]{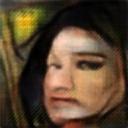}}}
\subfigure[\cite{zhu2017unpaired}]{\label{fig:cmp6f}\scalebox{1}[1]
{\includegraphics[width=0.118\linewidth]{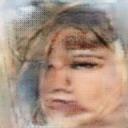}}}
\subfigure[\cite{Shiri2017FaceD}]{\label{fig:cmp6g}\scalebox{1}[1]
{\includegraphics[width=0.118\linewidth]{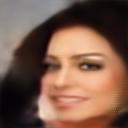}}}
\subfigure[Ours]{\label{fig:cmp6h}\scalebox{1}[1]
{\includegraphics[width=0.118\linewidth]{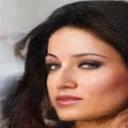}}}
\hfill
\vspace{-0.1cm}
\caption{Qualitative comparisons of the state-of-the-art methods. (a) Input portraits (from the test dataset) including seen and unseen styles. (b) Ground truth face images. (c) Gatys \emph{et al.}'s method~\cite{gatys2016image}. (d) Johnson \emph{et al.}'s method~\cite{johnson2016perceptual}. (e) Isola \emph{et al.}'s method~\cite{isola2016image} (pix2pix). (f) Zhu \emph{et al.}'s method~\cite{zhu2017unpaired} (CycleGAN). (g) Shiri \emph{et al.}'s method~\cite{Shiri2017FaceD} (h) Our method.}
\label{fig:cmp6}
\vspace{-0.3cm}
\end{figure*}
\subsection{Quantitative Evaluation}

\noindent{\textbf{Face Reconstruction Analysis. }}
To evaluate the reconstruction performance, we measure the average Peak Signal to Noise Ratio (PSNR), Structural Similarity (SSIM)~\cite{wang2004image} and Feature Similarity (FSIM)~\cite{zhang2011fsim} on the entire test dataset. Table~\ref{tab1} indicates that our IFRP method achieves superior quantitative results in comparison to other methods on both seen and unseen styles. Moreover, we also evaluate different methods on sketch images from the CUFSF dataset as an unseen style without fine-tuning or retraining our network.


\begin{table*}\renewcommand{\arraystretch}{1}
\centering
\caption{Comparisons of PSNR, SSIM and FSIM on the entire test dataset.}
\scalebox{1}{
\begin{tabular}{>{\centering\arraybackslash}m{0.12\linewidth}|>{\centering\arraybackslash}m{0.06\linewidth}|>{\centering\arraybackslash}m{0.06\linewidth}|>{\centering\arraybackslash}m{0.06\linewidth}|>{\centering\arraybackslash}m{0.06\linewidth}|>{\centering\arraybackslash}m{0.06\linewidth}|>{\centering\arraybackslash}m{0.06\linewidth}|>{\centering\arraybackslash}m{0.06\linewidth}|>{\centering\arraybackslash}m{0.06\linewidth}|>{\centering\arraybackslash}m{0.06\linewidth}}
\hline
\multirow{2}{*}{Methods} & \multicolumn{3}{c|}{Seen Styles} & \multicolumn{3}{c|}{Unseen Styles} & \multicolumn{3}{c}{Unseen Sketches} \\
\cline{2-10}
& PSNR & SSIM & FSIM & PSNR & SSIM &  FSIM & PSNR & SSIM & FSIM\\
\hline
Gatys ~\cite{gatys2016image} & 20.18 & 0.57 & 0.73 & 20.25 & 0.57 & 0.66 & 19.93 & 0.55 & 0.67\\ 
Johnson~\cite{johnson2016perceptual} & 15.65 & 0.34 & 0.68 & 15.81 & 0.33 & 0.70 & 16.27 & 0.35 & 0.68 \\
MGAN~\cite{li2016precomputed} & 16.22 & 0.44 & 0.64 & 16.17 & 0.47 & 0.60 & 16.01 & 0.46 & 0.61 \\
pix2pix~\cite{isola2016image} & 20.82 & 0.59 & 0.80 & 18.90 & 0.54 & 0.67 & 19.01 & 0.55 & 0.66\\
CycleGAN~\cite{zhu2017unpaired} & 18.58 & 0.32 & 0.69 & 15.89 & 0.27 & 0.64 & 15.65 & 0.31 & 0.65\\
Shiri~\cite{Shiri2017FaceD} & 21.57 & 0.58 & 0.79 & 20.21 & 0.56 & 0.70 & 21.35 & 0.57 & 0.71\\
\hline
\bf{IFRP} & {\bf 26.08} & {\bf 0.72} & {\bf 0.88} & {\bf 24.83} & {\bf 0.68} & {\bf 0.84} & {\bf 24.89} & {\bf 0.68}& {\bf 0.83}\\
\hline
\end{tabular}}
\label{tab1}
\end{table*}

\begin{table*}\renewcommand{\arraystretch}{1}
\centering
\caption{Comparisons of FRR and FCR on the entire test dataset.}
\begin{tabular}{>{\centering\arraybackslash}m{0.13\linewidth}|>{\centering\arraybackslash}m{0.13\linewidth}|>{\centering\arraybackslash}m{0.13\linewidth}|>{\centering\arraybackslash}m{0.13\linewidth}|>{\centering\arraybackslash}m{0.12\linewidth}}
\hline
\multirow{2}{*}{Methods} & \multicolumn{3}{c|}{FRR} & \multirow{2}{*}{FCR} \\
\cline{2-4}
& Seen Styles & Unseen Styles & Unseen Sketch & \\
\hline
Gatys ~\cite{gatys2016image}  & 64.67\%  & 62.28\% & 68.36\% & 72.89\%\\ 
Johnson~\cite{johnson2016perceptual}  & 50.54\% &  38.87\% & 40.27\% & 44.99\% \\
MGAN~\cite{li2016precomputed}  & 26.97\% & 22.52\% & 24.99\% & 38.24\% \\
pix2pix~\cite{isola2016image} & 75.13\%  & 59.98\% & 66.63\% & 87.73\%  \\
CycleGAN~\cite{zhu2017unpaired} &  25.07\% &  25.68\% & 26.70\% & 24.97\% \\
Shiri~\cite{Shiri2017FaceD} & 84.51\%  & 75.32\% & 76.44\% & 89.09\%\\
\hline
{\bf IFRP} & {\bf 90.93\%} &  {\bf 84.92\%} & {\bf 89.05\%} & {\bf 92.06\%} \\
\hline
\end{tabular}
\label{tab2}
\end{table*}

\vspace{0.05cm}
\noindent\textbf{Face Retrieval Analysis.} 
%
Below we demonstrate that the faces recovered by our method are highly consistent with their ground truth counterparts.
To this end, we run a face recognition algorithm~\cite{parkhi2015deep} on our test dataset for both seen and unseen styles. For each investigated method, we set 1K recovered faces from one style as a query dataset and then we set 1K of ground truth faces as a search dataset. 
We apply \cite{parkhi2015deep} to quantify whether the correct person is retrieved within the top-5 matched images. Then an average retrieval score is obtained. We repeat this procedure for every style and then obtain the average Face Retrieval Ratio (FRR) by averaging all scores from the seen and unseen styles, respectively. 
As indicated in Table~\ref{tab2}, our IFRP network outperforms the other methods across all the styles. Even for the unseen styles, our method can still retain most identity-preserving features, making the destylized results similar to the ground truth faces.
Moreover, we also run an experiment on hand-drawn sketches of the CUFSF dataset used as an unseen style. The FRR scores are higher compared to results on other styles as facial components are easier to extract from sketches/their contours. 
Despite our method is not dedicated to face retrieval, we compare it to \cite{zhang2011coupled}. To challenge our method, we retrain our network on sketches. To this end, we recovered faces from sketches (the CUFSF dataset) and performed face identification that yielded $\sim$91\% Verification Rate (VR) at FAR=0.1\%. This outperforms photo-synthesizing approach MRF+LE \cite{zhang2011coupled} which uses sketches for training and yields 43.66\% VR at FAR=0.1\%.

\vspace{0.05cm}
\noindent{\textbf{Consistency Analysis w.r.t. Various Styles.}}
As shown in Figures~\ref{fig:cmp1i} and \ref{fig:cmp2i}, our network recovers photorealistic face images from various stylized portraits of the same person. Note that recovered faces resemble each other. It indicates that our network is robust to different styles.

In order to demonstrate the robustness of our network to different styles quantitatively, we study the consistency of faces recovered from different styles.
First, we choose 1K face images destylized from a single style. For each destylized face, we search its top-5 most similar face images in a group of face images destylized from portraits in remaining styles. If the same person is retrieved within the top-5 candidates, we record this as a hit. Then an average hit number for a given style is obtained. We repeat the same procedure for each of the other 7 styles, and then we calculate the average hit number, denoted as Face Consistency Ratio (FCR). Note that the probability of one hit by chance is 0.5$\%$.
Table~\ref{tab2} shows average FCR scores on the test dataset for each method. The FCR scores indicate that our IFRP method produces the most consistent destylized face images across different styles. This also implies that our SRN can extract facial features irrespective of image styles.

\begin{table}\renewcommand{\arraystretch}{1}
\centering
\caption{Quantitative comparisons of the impact of each of our losses.}
\begin{tabular}{>{\centering\arraybackslash}m{0.38\linewidth}|>{\centering\arraybackslash}m{0.06\linewidth}|>{\centering\arraybackslash}m{0.06\linewidth}|>{\centering\arraybackslash}m{0.06\linewidth}|>{\centering\arraybackslash}m{0.04\linewidth}}
\hline
\multirow{2}{*}{Loss Function} & \multicolumn{2}{c|}{Seen Styles} & \multicolumn{2}{c}{Unseen Styles}\\
\cline{2-5}
& SSIM & FSIM & SSIM  & FSIM\\
\hline
$\mathcal{L}_{\rm pix}$ & 0.60 & 0.72 & 0.54 & 0.65\\
$\mathcal{L}_{pix}$ + $\mathcal{L}_{\rm dis}$ & 0.62 & 0.75 & 0.58 & 0.72\\
\bf{IFRP} ($\mathcal{L}_{\rm pix}$+$\mathcal{L}_{\rm dis}$+$\mathcal{L}_{id}$) & {\bf 0.72} & {\bf 0.88}  & {\bf 0.68} & {\bf 0.84}\\
\hline
\end{tabular}
\label{tab10}
\end{table}

\subsection{Impact of Different Losses on Performance.}
Below we discuss the  impact of our losses on the visual results shown in Figure~\ref{fig:DiscEffect} and we present corresponding quantitative evaluations in Table~\ref{tab10}.
Figure~\ref{fig:DiscEffect} shows that employing only the pixel-wise loss $\mathcal{L}_{\rm pix}$ leads to the visual recovery which suffers from severe blur, as $\mathcal{L}_{\rm pix}$ loss acts on the intensity similarity only. 
To avoid generating overly smooth results, the discriminative loss $\mathcal{L}_{\rm dis}$ is employed by us in our network. Similar to findings of our previous work~\cite{Shiri2017FaceD}, the discriminative loss encourages the generated faces to be realistic, thus it improves the final results qualitatively and quantitatively. The weight/impact of the discriminative loss is chosen experimentally with value of $10^{-2}$ being a good compromise between excessively smooth and sharp results. However, due to the lack of the guidance of high-level semantic information (parts of such information are locally lost in stylized portraits), the network with the pixel-wise and discriminative losses still generates artifacts to mimic facial details. 
As shown in Figure~\ref{fig:DisD} (top), the network still generates ambiguous results such as gender reversal or mismatched hair color. By employing all three losses together, that is, the identity-preserving loss $\mathcal{L}_{id}$, discriminative loss $\mathcal{L}_{\rm dis}$ and the pixel-wise loss $\mathcal{L}_{\rm pix}$, our network attains the best visual recovery. The same findings are confirmed by the quantitative results in Table ~\ref{tab10}.

\subsection{Ablation Study of the Proposed Architecture.}
We perform the ablation study of different components of the proposed IFRP architecture and present visual results in Figure~\ref{fig:ablation}. In order to demonstrate the contribution of each component to the quantitative results, we also show the quantitative results of our network in Table~\ref{tab11}. When only employing a standard autoencoder with a stack of convolutional layers followed by a series of deconvolutional layers, the visual results suffer from blurriness and artifacts as shown in Figure~\ref{fig:ablationC}. The network generates misguided results such as wrong hair texture, lack of lipstick, wrong lip expression, \etc 
To avoid generating overly smooth results, the skip connections between two top layers are applied in our network. By employing residual blocks between skip connection of top two layers, our network is able to achieve the best results qualitatively and quantitatively. In this manner, we put emphasis on  high-level semantic information.

\begin{table}\renewcommand{\arraystretch}{1}
\centering
\caption{Quantitative comparisons of the impact of various IFRP network components.}
\begin{tabular}{>{\raggedright\arraybackslash}m{0.463\linewidth}|>{\centering\arraybackslash}m{0.06\linewidth}|>{\centering\arraybackslash}m{0.06\linewidth}|>{\centering\arraybackslash}m{0.06\linewidth}|>{\centering\arraybackslash}m{0.02\linewidth}}
\hline
\multirow{2}{*}{SRN Architecture} & \multicolumn{2}{c|}{Seen Styles} & \multicolumn{2}{c}{Unseen Styles} \\
\cline{2-5}
& SSIM & FSIM & SSIM  & FSIM \\
\hline
Standard Autoencoder & 0.65 & 0.84 & 0.62 & 0.80\\\vspace{1mm}
U-net Autoencoder& 0.65 & 0.87 & 0.61 & 0.78\\\vspace{1mm}
Top 2-layer skip conn. & 0.66 & 0.86 & 0.63 & 0.82 \\\vspace{1mm}
\bf{IFRP:}\small{ 2-layer skip conn.+Res. blocks}& {\bf 0.72} & {\bf 0.88}  & {\bf 0.68} & {\bf 0.84} \\
\hline
\end{tabular}
\label{tab11}
\end{table}

\begin{table}\renewcommand{\arraystretch}{1}
\centering
\caption{SSIM as the function of the number of in-plain rotation-based augmentations of SF images used during training.}
\begin{tabular}{>{\centering\arraybackslash}m{0.60\linewidth}|>{\centering\arraybackslash}m{0.10\linewidth}|>{\centering\arraybackslash}m{0.10\linewidth}}
\hline
\centering
Rotation Angles (degrees) & Without STNs & With STN \\
\hline
-30, -20, -15, -10, -5, 0, 5, 10, 15, 20, 30 & 0.64 & 0.66 \\
-30, -15, 0, 15, 30 & 0.64 & 0.65\\
\hline
\end{tabular}
\label{tab-degrees}
\end{table}

\begin{figure*}
\centering
\includegraphics[width=0.107\linewidth]{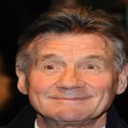}
\includegraphics[width=0.107\linewidth]{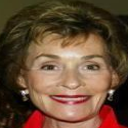}
\includegraphics[width=0.107\linewidth]{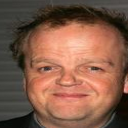}
\includegraphics[width=0.107\linewidth]{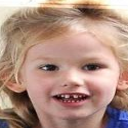}
\includegraphics[width=0.107\linewidth]{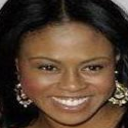}
\includegraphics[width=0.107\linewidth]{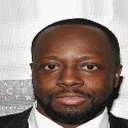}
\includegraphics[width=0.107\linewidth]{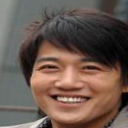}
\includegraphics[width=0.107\linewidth]{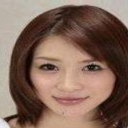}
\includegraphics[width=0.107\linewidth]{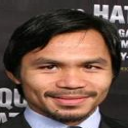}\\\vspace{0.7mm}
\includegraphics[width=0.107\linewidth]{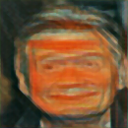}
\includegraphics[width=0.107\linewidth]{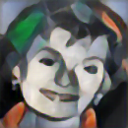}
\includegraphics[width=0.107\linewidth]{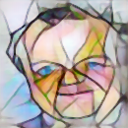}
\includegraphics[width=0.107\linewidth]{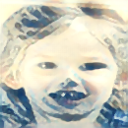}
\includegraphics[width=0.107\linewidth]{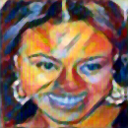}
\includegraphics[width=0.107\linewidth]{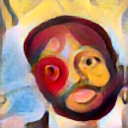}
\includegraphics[width=0.107\linewidth]{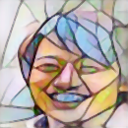}
\includegraphics[width=0.107\linewidth]{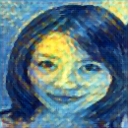}
\includegraphics[width=0.107\linewidth]{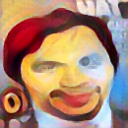}\\\vspace{-1.2mm}
\subfigure[]{\label{fig:eth1}\scalebox{1}[1]{\includegraphics[width=0.107\linewidth]{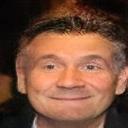}}}
\subfigure[]{\label{fig:eth2}\scalebox{1}[1]{\includegraphics[width=0.107\linewidth]{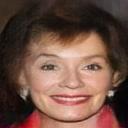}}}
\subfigure[]{\label{fig:eth3}\scalebox{1}[1]{\includegraphics[width=0.107\linewidth]{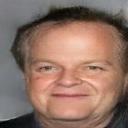}}}
\subfigure[]{\label{fig:eth4}\scalebox{1}[1]{\includegraphics[width=0.107\linewidth]{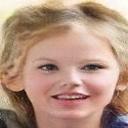}}}
\subfigure[]{\label{fig:eth5}\scalebox{1}[1]{\includegraphics[width=0.107\linewidth]{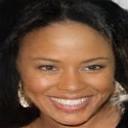}}}
\subfigure[]{\label{fig:eth6}\scalebox{1}[1]{\includegraphics[width=0.107\linewidth]{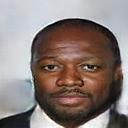}}}
\subfigure[]{\label{fig:eth7}\scalebox{1}[1]{\includegraphics[width=0.107\linewidth]{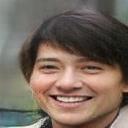}}}
\subfigure[]{\label{fig:eth8}\scalebox{1}[1]{\includegraphics[width=0.107\linewidth]{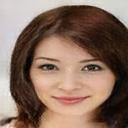}}}
\subfigure[]{\label{fig:eth9}\scalebox{1}[1]{\includegraphics[width=0.107\linewidth]{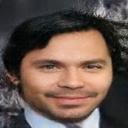}}}
\caption{Examples of images recovered from portraits of different ethnicity and age groups. First row: RF ground truth faces. Second row: unaligned input portraits. Third row: our recovery results. (a-d) Faces of very old or young subjects. (e-f) Faces of dark skinned subjects. (g-i) Faces of Asian subjects.}
\label{fig:ethnicity}
\end{figure*}
\subsection{Robustness of IFRP w.r.t. Ethnicity and Age}
We note that the numbers of images of children, old people and young adults in the CelebA dataset are unbalanced \eg, there are more images of young adults than children and old people. Moreover, the number of images of people of white complexion is larger compared to those of dark skin tones. The number of Asian faces is also limited in the CelebA dataset. Unfortunately, these factors make our synthesized dataset unbalanced. However, due to the identity-preserving loss we use, our network can cope with faces of different nationalities, skin tones and ages reasonably well. Figure~\ref{fig:ethnicity} shows the visual results obtained by our network given faces of various ethnicity and age. Our results are consistent with the ground truth face images. However, some age-related facial features such as children's missing teeth in Figure \ref{fig:eth4} (bottom) are especially hard to recover faithfully as CelebA does not feature celebrities with missing teeth \etc.

\subsection{Robustness of IFRP w.r.t. Misalignments}
\begin{figure*}
\centering
\includegraphics[width=0.12\linewidth]{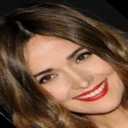}
\includegraphics[width=0.12\linewidth]{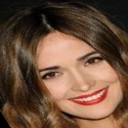}
\includegraphics[width=0.12\linewidth]{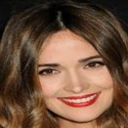}
\includegraphics[width=0.12\linewidth]{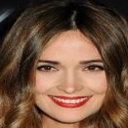}
\includegraphics[width=0.12\linewidth]{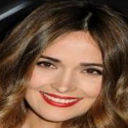}
\includegraphics[width=0.12\linewidth]{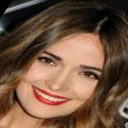}
\includegraphics[width=0.12\linewidth]{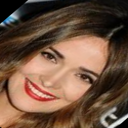}\\\vspace{0.1em}
\includegraphics[width=0.12\linewidth]{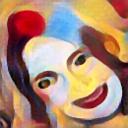}
\includegraphics[width=0.12\linewidth]{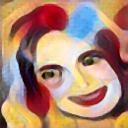}
\includegraphics[width=0.12\linewidth]{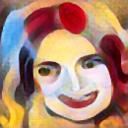}
\includegraphics[width=0.12\linewidth]{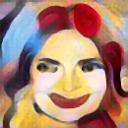}
\includegraphics[width=0.12\linewidth]{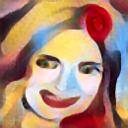}
\includegraphics[width=0.12\linewidth]{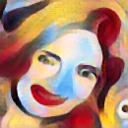}
\includegraphics[width=0.12\linewidth]{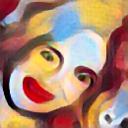}\\\vspace{-0.45em}
\subfigure[-45]{\label{fig:-45}{\includegraphics[width=0.12\linewidth]{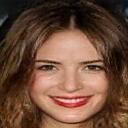}}}
\subfigure[-30]{\label{fig:-30}{\includegraphics[width=0.12\linewidth]{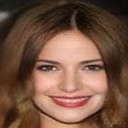}}}
\subfigure[-15]{\label{fig:-15}{\includegraphics[width=0.12\linewidth]{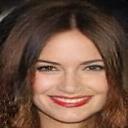}}}
\subfigure[0]{\label{fig:0}{\includegraphics[width=0.12\linewidth]{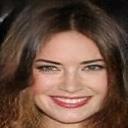}}}
\subfigure[+15]{\label{fig:15}{\includegraphics[width=0.12\linewidth]{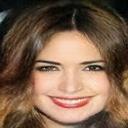}}} 
\subfigure[+30]{\label{fig:15b}{\includegraphics[width=0.12\linewidth]{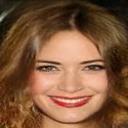}}} 
\subfigure[+45]{\label{fig:45}{\includegraphics[width=0.12\linewidth]{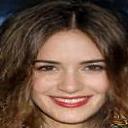}}} 
\caption{The effect of STN layers on recovery from unaligned rotated portraits ([-45; 45] degrees range). First row: the ground truth face image. Second row: unaligned rotated portraits using \emph{Candy} style. Last row: our aligned results.}
\label{fig:STN}
\end{figure*}

\begin{figure}
\begin{center}
\includegraphics[width = 1\linewidth]{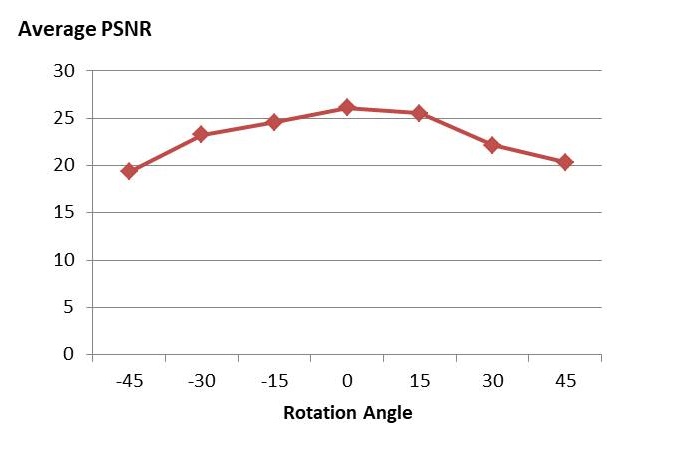}
\caption{Performance of our IFRP w.r.t. the rotation angle of faces.}
\label{fig:Angle_PSNR}
\end{center}
\end{figure}

\begin{figure}
\centering
\includegraphics[width=0.18\linewidth]{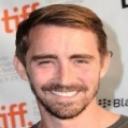}
\includegraphics[width=0.18\linewidth]{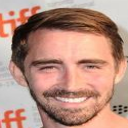}
\includegraphics[width=0.18\linewidth]{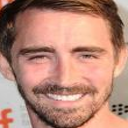}\\\vspace{0.1em}
\includegraphics[width=0.18\linewidth]{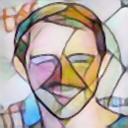}
\includegraphics[width=0.18\linewidth]{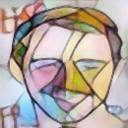}
\includegraphics[width=0.18\linewidth]{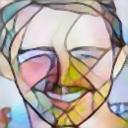}\\\vspace{-0.45em}
\subfigure[0.7x]{\label{fig:0.7}\scalebox{1}[1]{\includegraphics[width=0.18\linewidth]{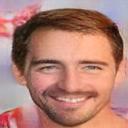}}}
\subfigure[1x]{\label{fig:1}\scalebox{1}[1]{\includegraphics[width=0.18\linewidth]{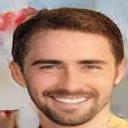}}}
\subfigure[1.3x]{\label{fig:1.3}\scalebox{1}[1]{\includegraphics[width=0.18\linewidth]{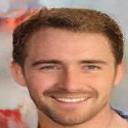}}}
\caption{The effect of STN layers on portrait scaling. First row: the ground truth face image. Second row: unaligned stylized faces using \emph{ Mosaic} style. Last row: our aligned results.}
\label{fig:Scale}
\end{figure}

\begin{figure}
\begin{center}
\includegraphics[width = 1\linewidth]{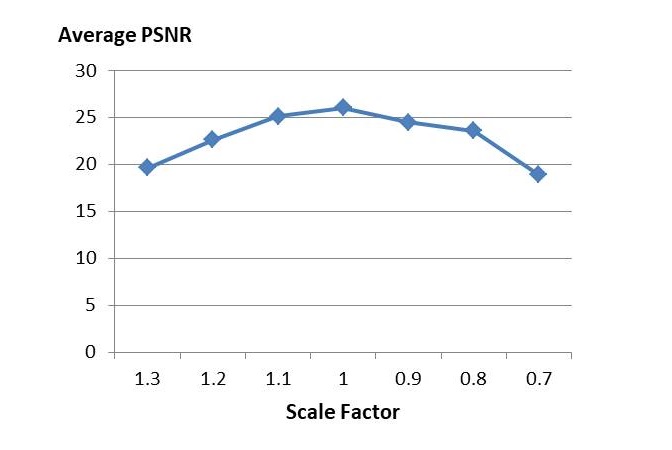}
\caption{Performance of our IFRP w.r.t. the scale of faces.}
\label{fig:Scale_PSNR}
\end{center}
\end{figure}

Below we conduct some  qualitative and quantitative experiments to show the robustness of our network to misalignments. Figure~\ref{fig:STN} shows the visual results of our network on faces rotated within range [-45; 45] degrees. Thanks to STN layers, our network is able to recover photorealistic faces even from portraits rotated by -45 or +45 degrees. Figure~\ref{fig:Angle_PSNR} shows PSNR of our network as a function of the rotation angle. Moreover, our network is also robust to scaling of portraits. Figure~\ref{fig:Scale} shows the successfully recovered faces from portraits containing faces captured at different scales. Figure~\ref{fig:Scale_PSNR} shows PSNR of our network as a function of the scale factor.

Moreover, Table \ref{tab-degrees} shows SSIM  scores for a single-style training with only in-plane rotations of SF used during training. The table shows that using STN layers benefits results. However, using STNs is only a discretionary choice.

\begin{figure}[t]
\centering
\includegraphics[width=0.16\linewidth]{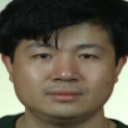}
\includegraphics[width=0.16\linewidth]{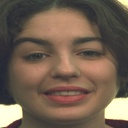}
\includegraphics[width=0.16\linewidth]{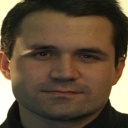}
\includegraphics[width=0.16\linewidth]{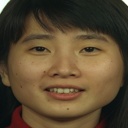}\\
\includegraphics[width=0.16\linewidth]{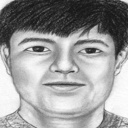}
\includegraphics[width=0.16\linewidth]{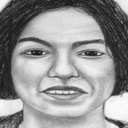}
\includegraphics[width=0.16\linewidth]{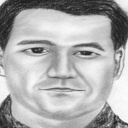}
\includegraphics[width=0.16\linewidth]{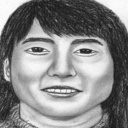}\\
\includegraphics[width=0.16\linewidth]{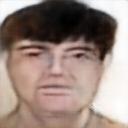}
\includegraphics[width=0.16\linewidth]{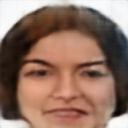}
\includegraphics[width=0.16\linewidth]{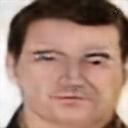}
\includegraphics[width=0.16\linewidth]{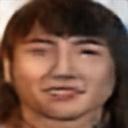}\\
\includegraphics[width=0.16\linewidth]{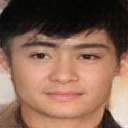}
\includegraphics[width=0.16\linewidth]{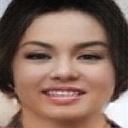}
\includegraphics[width=0.16\linewidth]{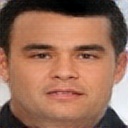}
\includegraphics[width=0.16\linewidth]{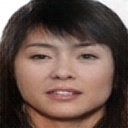}\\
\caption{Recovering photorealistic faces from hand-drawn sketches from the FERET dataset. First row: ground truth face images. Second row: sketches. Third row: results of Wang \emph{et al.}'s method~\cite{wang2018high}. Bottom row: our results.}
\label{fig:sketch}
\end{figure}


\subsection{User Experience Study}
\vspace{-0.2cm}
As human perception is sensitive to the slightest imperfections and artifacts of faces, we conducted a user study to verify if subjects find convincing our recovered results.

\begin{figure}
\begin{center}
\includegraphics[trim=0 20 0 35, clip=true,width = 1\linewidth]{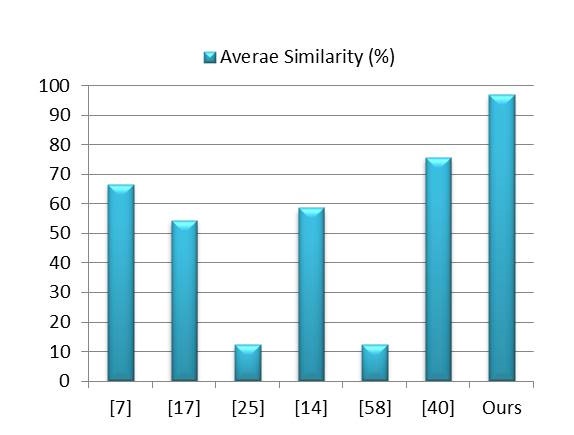}\vspace{-0.1cm}\\
\begin{tabular}{c c c c c c c}
\hspace{0.85cm}\cite{gatys2016image} &
\hspace{0.03cm}\cite{johnson2016perceptual} &
\hspace{0.03cm}\cite{li2016precomputed} &
\hspace{0.03cm}\cite{isola2016image} &
\hspace{0.03cm}\cite{zhu2017unpaired} &
\hspace{0.03cm}\cite{Shiri2017FaceD} &
Ours
\end{tabular}
\caption{User study. Results comparing our IFRP and other state-of-the-art methods. Vertical axis is the percentages of favorable user votes.}
\label{fig:BarChart}
\end{center}
\end{figure}

Our evaluation dataset contains faces recovered from 20 stylized portraits by the state-of-the-art methods as well as our IFRP method (see an example in Figure \ref{fig:cmp6}). We chose a diverse subset of portraits in terms of race, gender, age, hair style, skin color, make up, \etc Our study included 25 subjects (graduate students).
For each portrait, the Ground truth face and seven images (the faces recovered by \cite{gatys2016image,johnson2016perceptual,li2016precomputed,isola2016image,zhu2017unpaired,Shiri2017FaceD} as well as our method) were shown in random order side-by-side on high-quality color printouts. 
The subjects were asked to rate the printouts according to the visual quality and perceived fidelity of identity with respect to the corresponding ground truth images. 
Figure~\ref{fig:BarChart} summarizes the average scores of this study. For all portraits, our results are rated higher than other state-of-the-art methods. The subjects rated higher the printouts which preserve the subjects' identities better and contain no visible artifacts. As this simple user study shows, our results are  favored by the users as they find faces recovered by our algorithm to be the closest to the original images. This study is consistent with our numerical evaluations.

\subsection{Destylizing Authentic Paintings and Sketches}
\begin{figure}
\centering
\includegraphics[width=0.16\linewidth]{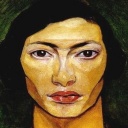}
\includegraphics[width=0.16\linewidth]{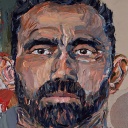}
\includegraphics[width=0.16\linewidth]{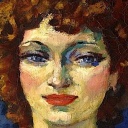}
\includegraphics[width=0.16\linewidth]{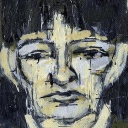}
\includegraphics[width=0.16\linewidth]{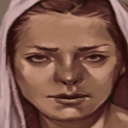}\\\vspace{0.2em}
\includegraphics[width=0.16\linewidth]{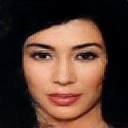}
\includegraphics[width=0.16\linewidth]{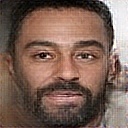}
\includegraphics[width=0.16\linewidth]{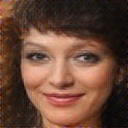}
\includegraphics[width=0.16\linewidth]{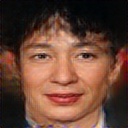}
\includegraphics[width=0.16\linewidth]{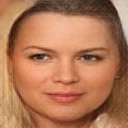}
\caption{Recovery results for the authentic unaligned paintings. Top row: the original portraits from art galleries. Bottom row: our results.}
\label{fig:Orig}
\end{figure}

Below we demonstrate that our method is not restricted to the recovery of faces from computer-generated stylized portraits but it can also work with real paintings, sketches and unknown styles. To verify this assertion, we choose a few of paintings from art galleries such as Archibald~\cite{archibald}. Next, we crop face regions from the scanned images and use them as our test images. 
Figure \ref{fig:Orig} shows that our method can efficiently recover photorealistic face images. This indicates that our method is not limited to the synthesized data and it does not require an alignment procedure beforehand.

We also conduct an experiment on hand-drawn sketches from the FERET dataset~\cite{phillips1998feret}. We compare our results with one of the most recent sketch-to-face methods.
Method~\cite{wang2018high} works with sketches only (\emph{c.f.} complex stylized faces) and requires landmarks to perform the face alignment (\emph{c.f.} our method which does not need any face alignment due to STN layers). Note that \cite{wang2018high} uses CycleGAN with multipatch-based discriminators to generate sketches/photos. Figure \ref{fig:sketch} shows the comparison of our method with method~\cite{wang2018high} which is not fully supervised and tends to produce artifacts. 
In contrast, our method can efficiently recover photorealistic face images from sketches and it results in fewer artifacts  due to the identity-preserving loss.

\begin{figure}[t]
\centering
\includegraphics[width=0.17\linewidth]{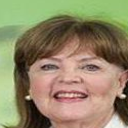}
\includegraphics[width=0.17\linewidth]{figs/201299.png}
\includegraphics[width=0.17\linewidth]{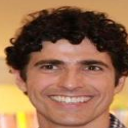}
\includegraphics[width=0.17\linewidth]{figs/199685.png}
\includegraphics[width=0.17\linewidth]{figs/199685.png}\\
\includegraphics[width=0.17\linewidth]{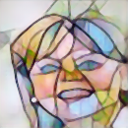}
\includegraphics[width=0.17\linewidth]{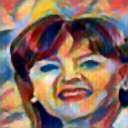}
\includegraphics[width=0.17\linewidth]{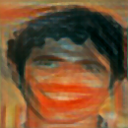}
\includegraphics[width=0.17\linewidth]{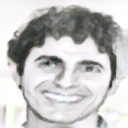}
\includegraphics[width=0.17\linewidth]{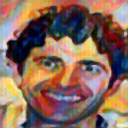}\\
\subfigure[Seen]{\label{fig:failure1}{\includegraphics[width=0.17\linewidth]{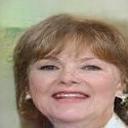}}}
\subfigure[Unseen]{\label{fig:failure2}{\includegraphics[width=0.17\linewidth]{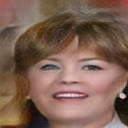}}}
\subfigure[Seen]{\label{fig:failure3}{\includegraphics[width=0.17\linewidth]{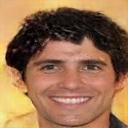}}}
\subfigure[Unseen]{\label{fig:failure4}{\includegraphics[width=0.17\linewidth]{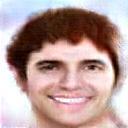}}}
\subfigure[Unseen]{\label{fig:failure4b}{\includegraphics[width=0.17\linewidth]{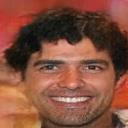}}}
\caption{Limitations. Top row: ground truth face images. Middle row: unaligned stylized face images. Bottom row: our results.}
\label{fig:Failure}
\end{figure}

\begin{figure}[t]
\centering
\includegraphics[width=0.24\linewidth]{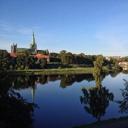}
\includegraphics[width=0.24\linewidth]{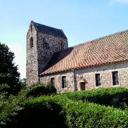}
\includegraphics[width=0.24\linewidth]{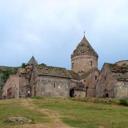}
\includegraphics[width=0.24\linewidth]{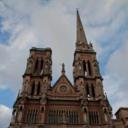}\\
\includegraphics[width=0.24\linewidth]{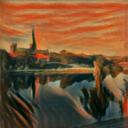}
\includegraphics[width=0.24\linewidth]{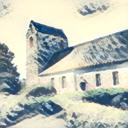}
\includegraphics[width=0.24\linewidth]{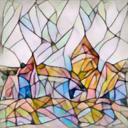}
\includegraphics[width=0.24\linewidth]{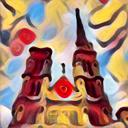}\\
\includegraphics[width=0.24\linewidth]{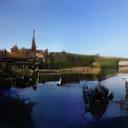}
\includegraphics[width=0.24\linewidth]{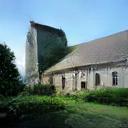}
\includegraphics[width=0.24\linewidth]{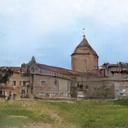}
\includegraphics[width=0.24\linewidth]{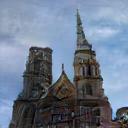}\\
\caption{Results of our IPFR approach on the \emph{Church Outdoor} dataset~\cite{yu15lsun}. Top row: ground truth images. Middle row: stylized images. Bottom row: our results.}
\label{fig:church}
\end{figure}
\subsection{Image Recovery from Generic Artworks}
Below we conduct an experiment on the Church outdoor dataset to show that our network can recover photorealistic images from generic artworks (\emph{c.f.} portraits). Figure \ref{fig:church} demonstrates the ground truth images, stylized images and images recovered by our IFRP network, respectively. We note that the diversity of outdoor images is much larger than those of faces. Therefore, as expected, for a reliable training and recovery of generic scenes, big datasets are needed. Nonetheless, our recovered results are visually convincing.

\subsection{Limitations on Unseen Styles} 
 

We have noted that our network is able to recover peripheral non-facial details for styles both seen and used during training. Figure~\ref{fig:Failure} shows that the background color and texture for seen styles (\emph{Mosaic} and \emph{Scream}) are fully recovered as the background information is encoded in the stylized images.
For styles (\emph{Composition VII} and \emph{Sketch}) unseen during training, our network hallucinated backgrounds inconsistent with the ground truth backgrounds. As expected in this sanity check, the recovered background colors and textures for unseen stylized portraits do not match the ground truth.

\section{Conclusions}

We have introduced a novel neural network for the face recovery from stylized portraits. Our method extracts features from a given unaligned stylized portrait and then recovers a photorealistic face image from these features. The SRN successfully learns a mapping from unaligned stylized faces to aligned photorealistic faces. Our identity-preserving loss further encourages our network to generate identity trustworthy faces. This makes our algorithm readily available for the use in face hallucination, recovery and recognition. We have shown that our approach can recover images of faces from portraits of unseen styles, real paintings and sketches. Lastly, our approach can also recover some generic scenes and objects. In the future, we intend to embed semantic information into our network to generate more consistent face images in terms of semantic details.

\begin{acknowledgements}
This work is supported by the Australian Research Council (ARC) grant DP150104645.
\end{acknowledgements}

\appendix


\section{Face Alignment: Spatial Transfer Networks (STN).}
\label{sec:a1}
As described in Section {\color{red}{3.1}}, we incorporate multiple STNs~\cite{jaderberg2015spatial} as intermediate layers to compensate for misalignments and in-plane rotations.  
The STN layers estimate the motion parameters of face images and warp them to a canonical view. 
Each STN contains  localization, grid generator and sampler modules. The localization module consists of several hidden layers to estimate the transformation parameters with respect to the canonical view. The grid generator module creates a sampling grid according to the estimated parameters. Finally, the sampler module maps the input feature maps into generated girds using the bilinear interpolation.
The architecture of our STN layers is detailed in Tables~\ref{tab3},~\ref{tab4},~\ref{tab5} and ~\ref{tab6}.

\section{Contributions of each Component in the IFRP Network.}
\begin{figure}[t]
\centering
\includegraphics[width=0.18\linewidth]{figs/199685_Com.png}
\includegraphics[width=0.18\linewidth]{figs/199685.png}
\includegraphics[width=0.18\linewidth]{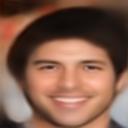}
\includegraphics[width=0.18\linewidth]{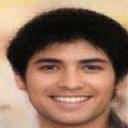}
\includegraphics[width=0.18\linewidth]{figs/199685_Com_D.jpg}\\
\includegraphics[width=0.18\linewidth]{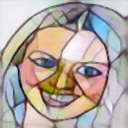}
\includegraphics[width=0.18\linewidth]{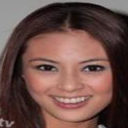}
\includegraphics[width=0.18\linewidth]{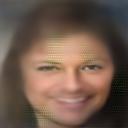}
\includegraphics[width=0.18\linewidth]{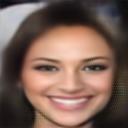}
\includegraphics[width=0.18\linewidth]{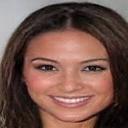}\\
\subfigure[]{\label{fig:A}\scalebox{1}[1]{\includegraphics[width=0.18\linewidth]{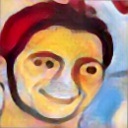}}}
\subfigure[]{\label{fig:B}\scalebox{1}[1]{\includegraphics[width=0.18\linewidth]{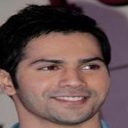}}}
\subfigure[]{\label{fig:C}\scalebox{1}[1]{\includegraphics[width=0.18\linewidth]{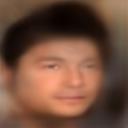}}}
\subfigure[]{\label{fig:D}\scalebox{1}[1]{\includegraphics[width=0.18\linewidth]{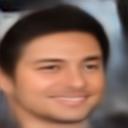}}}
\subfigure[]{\label{fig:E}\scalebox{1}[1]{\includegraphics[width=0.18\linewidth]{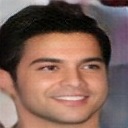}}}
\caption{More results showing the contribution of each loss function in our IFRP network. (a) Ground truth face images. (b) Input unaligned portraits from test dataset. (c) Recovered face images; the pixel-wise loss was used in training (no DN or identity-preserving losses). (d) Recovered face images; the pixel-wise loss and discriminative loss were used (no identity-preserving loss). (e) Our final results with the pixel-wise loss, discriminative loss and identity-preserving loss used during training.}
\label{fig:contribution}  
\end{figure}

In Section {\color{red}3}, we described the impact of the $\ell_2$ loss, the adversarial loss and the identity-preserving loss on the face recovery from portraits. 
Figure \ref{fig:contribution} further shows the contribution of each loss function to the final results.

\section{Visual Comparison with the State of the Art.}
\begin{figure*}[t]
\begin{minipage}{0.089\linewidth}
\centering
\subfigure[RF]{\label{fig:cmp3rf}\scalebox{1}[1]
{\includegraphics[width=1.18\linewidth]{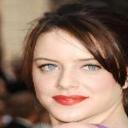}}}
\end{minipage}
\begin{minipage}{0.89\linewidth}
\centering
\hspace{0.5em}\includegraphics[width=0.118\linewidth]{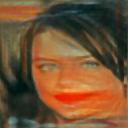}
\includegraphics[width=0.118\linewidth]{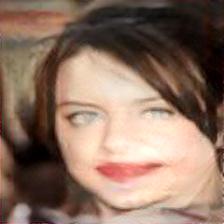}
\includegraphics[width=0.118\linewidth]{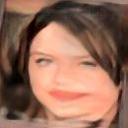}
\includegraphics[width=0.118\linewidth]{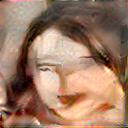}
\includegraphics[width=0.118\linewidth]{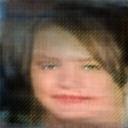}
\includegraphics[width=0.118\linewidth]{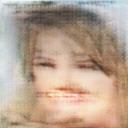}
\includegraphics[width=0.118\linewidth]{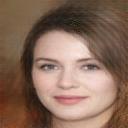}
\includegraphics[width=0.118\linewidth]{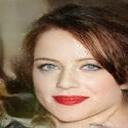}\\\vspace{0.1em}
\hspace{0.5em}\includegraphics[width=0.118\linewidth]{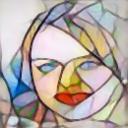}
\includegraphics[width=0.118\linewidth]{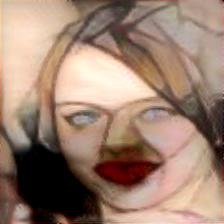}
\includegraphics[width=0.118\linewidth]{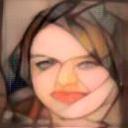}
\includegraphics[width=0.118\linewidth]{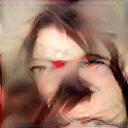}
\includegraphics[width=0.118\linewidth]{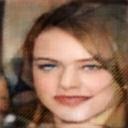}
\includegraphics[width=0.118\linewidth]{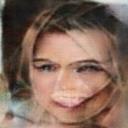}
\includegraphics[width=0.118\linewidth]{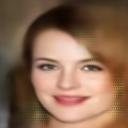}
\includegraphics[width=0.118\linewidth]{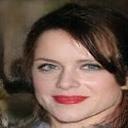}\\\vspace{0.1em}
\hspace{0.5em}\includegraphics[width=0.118\linewidth]{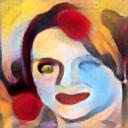}
\includegraphics[width=0.118\linewidth]{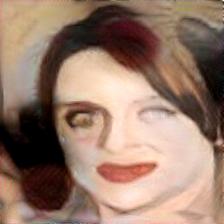}
\includegraphics[width=0.118\linewidth]{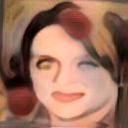}
\includegraphics[width=0.118\linewidth]{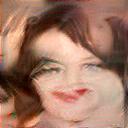}
\includegraphics[width=0.118\linewidth]{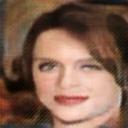}
\includegraphics[width=0.118\linewidth]{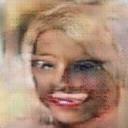}
\includegraphics[width=0.118\linewidth]{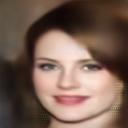}
\includegraphics[width=0.118\linewidth]{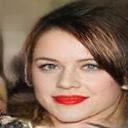}\\\vspace{0.1em}
\hspace{0.5em}\includegraphics[width=0.118\linewidth]{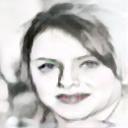}
\includegraphics[width=0.118\linewidth]{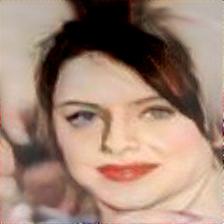}
\includegraphics[width=0.118\linewidth]{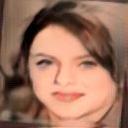}
\includegraphics[width=0.118\linewidth]{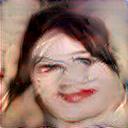}
\includegraphics[width=0.118\linewidth]{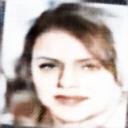}
\includegraphics[width=0.118\linewidth]{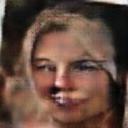}
\includegraphics[width=0.118\linewidth]{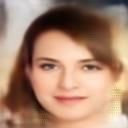}
\includegraphics[width=0.118\linewidth]{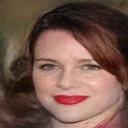}\\\vspace{0.1em}
\hspace{0.5em}\includegraphics[width=0.118\linewidth]{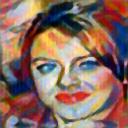}
\includegraphics[width=0.118\linewidth]{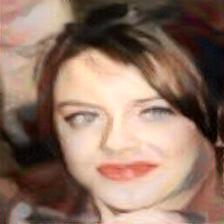}
\includegraphics[width=0.118\linewidth]{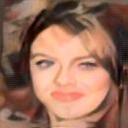}
\includegraphics[width=0.118\linewidth]{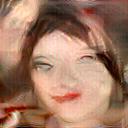}
\includegraphics[width=0.118\linewidth]{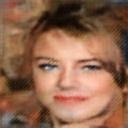}
\includegraphics[width=0.118\linewidth]{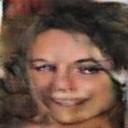}
\includegraphics[width=0.118\linewidth]{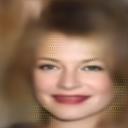}
\includegraphics[width=0.118\linewidth]{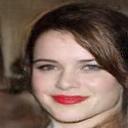}\\\vspace{0.1em}
\hspace{0.5em}\includegraphics[width=0.118\linewidth]{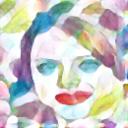}
\includegraphics[width=0.118\linewidth]{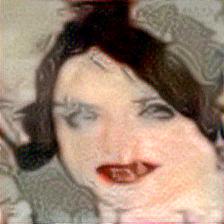}
\includegraphics[width=0.118\linewidth]{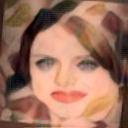}
\includegraphics[width=0.118\linewidth]{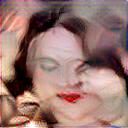}
\includegraphics[width=0.118\linewidth]{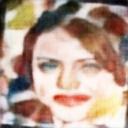}
\includegraphics[width=0.118\linewidth]{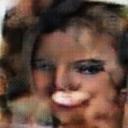}
\includegraphics[width=0.118\linewidth]{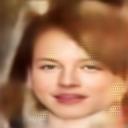}
\includegraphics[width=0.118\linewidth]{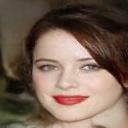}\\\vspace{0.1em}
\hspace{0.5em}\includegraphics[width=0.118\linewidth]{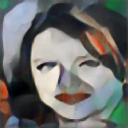}
\includegraphics[width=0.118\linewidth]{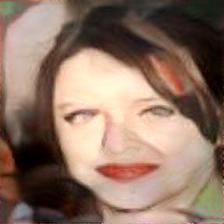}
\includegraphics[width=0.118\linewidth]{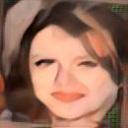}
\includegraphics[width=0.118\linewidth]{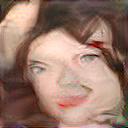}
\includegraphics[width=0.118\linewidth]{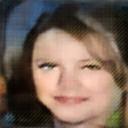}
\includegraphics[width=0.118\linewidth]{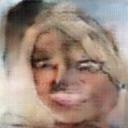}
\includegraphics[width=0.118\linewidth]{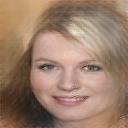}
\includegraphics[width=0.118\linewidth]{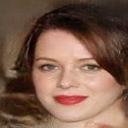}\\\vspace{-1.3mm}
\hspace{0.5em}\subfigure[SF]{\label{fig:cmp3b}\scalebox{1}[1]{\includegraphics[width=0.118\linewidth]{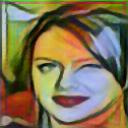}}}
\subfigure[\cite{gatys2016image}]{\label{fig:cmp3c}\scalebox{1}[1]
{\includegraphics[width=0.118\linewidth]{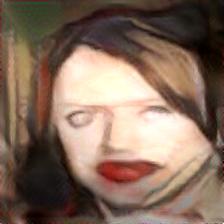}}}
\subfigure[\cite{johnson2016perceptual}]{\label{fig:cmp3d}\scalebox{1}[1]
{\includegraphics[width=0.118\linewidth]{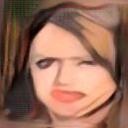}}}
\subfigure[\cite{li2016precomputed}]{\label{fig:cmp3e}\scalebox{1}[1]
{\includegraphics[width=0.118\linewidth]{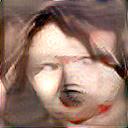}}}
\subfigure[\cite{isola2016image}]{\label{fig:cmp3f}\scalebox{1}[1]
{\includegraphics[width=0.118\linewidth]{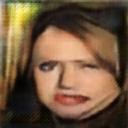}}}
\subfigure[\cite{zhu2017unpaired}]{\label{fig:cmp3g}\scalebox{1}[1]
{\includegraphics[width=0.118\linewidth]{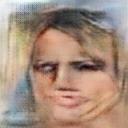}}}
\subfigure[\cite{Shiri2017FaceD}]{\label{fig:cmp3h}\scalebox{1}[1]
{\includegraphics[width=0.118\linewidth]{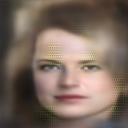}}}
\subfigure[Ours]{\label{fig:cmp3i}\scalebox{1}[1]
{\includegraphics[width=0.118\linewidth]{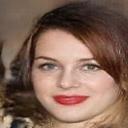}}}
\end{minipage}
\hfill
\vspace{-0.1cm}
\caption{Additional qualitative comparisons of the state-of-the-art methods. (a) Ground truth  face images. (b) Input portraits (from the test dataset) including the seen styles \emph{Scream, Mosaic} and \emph{Candy} as well as the unseen styles \emph{Sketch, Composition VII, Feathers, Udnie} and \emph{La Muse}. (c) Gatys \emph{et al.}'s method~\cite{gatys2016image}. (d) Johnson \emph{et al.}'s method~\cite{johnson2016perceptual}. (e) Li and Wand's method~\cite{li2016precomputed} (MGAN). (f) Isola \emph{et al.}'s method~\cite{isola2016image} (pix2pix). (g) Zhu \emph{et al.}'s method~\cite{zhu2017unpaired} (CycleGAN). (h) Shiri \emph{et al.}'s method~\cite{Shiri2017FaceD} (i) Our method.}
\label{fig:cmp3}
\end{figure*}

Below, we provide several additional results demonstrating the performance of our IFRP network compared to the state-of-art approaches (Figure \ref{fig:cmp3}).
\begin{table}[!ht]\renewcommand{\arraystretch}{1.2}
\vspace{-1.1em}
\caption{The STN1 architecture.}
\begin{center}
\begin{tabular}{>{\centering\arraybackslash}p{0.9\linewidth}}
\hline
STN1 \\
\hline
 Input: 64 x 64 x 32  \\
3 x 3 x 64 conv, relu, Max-pooling(2,2)  \\
3 x 3 x 128 conv, relu, Max-pooling(2,2) \\
3 x 3 x 256 conv, relu, Max-pooling(2,2)  \\
3 x 3 x 20 conv, relu, Max-pooling(2,2)  \\
3 x 3 x 20 conv, relu\\
fully connected (80,20), relu \\
fully connected (20,4) \\
\hline
\end{tabular}
\label{tab3}
\end{center}
\vspace{-1.1em}
\end{table}
\begin{table}[!ht]\renewcommand{\arraystretch}{1.2}
\vspace{-1.1em}
\caption{The STN2 architecture.}
\begin{center}
\begin{tabular}{>{\centering\arraybackslash}p{0.9\linewidth}}
\hline
STN2 \\
\hline
Input: 32 x 32 x 64  \\
3 x 3 x 128 conv, relu, Max-pooling(2,2) \\
3 x 3 x 256 conv, relu, Max-pooling(2,2)  \\
3 x 3 x 20 conv, relu, Max-pooling(2,2)  \\
3 x 3 x 20 conv, relu\\
fully connected (80,20), relu  \\
fully connected (20,4) \\
\hline
\end{tabular}
\label{tab4}
\end{center}
\vspace{-1.1em}
\end{table}
\begin{table}[!ht]\renewcommand{\arraystretch}{1.2}
\vspace{-1.1em}
\caption{The STN3 architecture.}
\begin{center}
\begin{tabular}{>{\centering\arraybackslash}p{0.9\linewidth}}
\hline
STN3 \\
\hline
Input: 16 x 16 x 128 \\
3 x 3 x 256 conv, relu, Max-pooling(2,2)  \\
3 x 3 x 20 conv, relu, Max-pooling(2,2)  \\
3 x 3 x 20 conv, relu\\
fully connected (80,20), relu  \\
fully connected (20,4)\\
\hline
\end{tabular}
\label{tab5}
\end{center}
\vspace{-1.1em}
\end{table}
\begin{table}[!ht]\renewcommand{\arraystretch}{1.2}
\vspace{-1.1em}
\caption{The STN4 architecture.}
\begin{center}
\begin{tabular}{>{\centering\arraybackslash}p{0.9\linewidth}}
\hline
 STN4\\
\hline
Input: 32 x 32 x 64\\
3 x 3 x 64 conv, relu, Max-pooling(2,2)\\
3 x 3 x 128 conv, relu, Max-pooling(2,2)  \\
3 x 3 x 256 conv, relu, Max-pooling(2,2)  \\
3 x 3 x 20 conv, relu\\
fully connected (80,20), relu \\
fully connected (20,4)\\
\hline
\end{tabular}
\label{tab6}
\end{center}
\vspace{-1.1em}
\end{table}

{\small
\bibliographystyle{spmpsci}
\bibliography{egbib}
}

\end{document}